\documentclass[sigconf]{acmart}

\usepackage{booktabs} 
\usepackage{comment}
\usepackage{multirow}
\usepackage{comment}
\usepackage{color}
\usepackage{bm}
\usepackage{url}

\setcopyright{rightsretained}

\newcommand{\bA}{{\mathbf A}}
\newcommand{\bB}{{\mathbf B}}
\newcommand{\bbR}{{\mathbb R}}

\newcommand{\calY}{{\mathcal Y}}
\newcommand{\calC}{{\mathcal C}}

\newcommand{\bW}{{\mathbf W}}
\newcommand{\bU}{{\mathbf U}}

\newcommand{\ind}[1]{\bm{1}_{\{#1\}}} 

\newcommand{\set}[1]{\left\{#1\right\}}
\newcommand{\abs}[1]{\left\lvert #1 \right\rvert}
\newcommand{\paren}[1]{\left(#1\right)}

\usepackage{subfigure}

\copyrightyear{2017} 
\acmYear{2017} 
\setcopyright{acmcopyright}
\acmConference{KDD '17}{August 13-17, 2017}{Halifax, NS, Canada}\acmPrice{15.00}\acmDOI{10.1145/3097983.3098166}
\acmISBN{978-1-4503-4887-4/17/08}
\begin{document}
\title{End-to-end Learning for Short Text Expansion}
\titlenote{Produces the permission block, and
  copyright information}
  

\author{Jian Tang$^1$, Yue Wang$^2$, Kai Zheng$^3$, Qiaozhu Mei$^{1,2}$}
\affiliation{%
  \institution{$^1$School of Information, University of Michigan}  
  \institution{$^2$Department of EECS, University of Michigan}
  \institution{$^3$Department of Informatics, University of California, Irvine}  
}
\email{jiant,raywang,qmei@umich.edu,\space \space kai.zheng@uci.edu}

\begin{abstract}






Effectively making sense of short texts is a critical task for many real world applications such as search engines, social media services, and recommender systems. The task is particularly challenging as a short text contains very sparse information, often too sparse for a machine learning algorithm to pick up useful signals. A common practice for analyzing short text is to first expand it with external information, which is usually harvested from a large collection of longer texts. In literature, short text expansion has been done with all kinds of heuristics. 
We propose an end-to-end solution that automatically learns how to expand short text to optimize a given learning task. A novel deep memory network is proposed to automatically find relevant information from a collection of longer documents and reformulate the short text through a gating mechanism. Using short text classification as a demonstrating task, we show that the deep memory network significantly outperforms classical text expansion methods with comprehensive experiments on real world data sets.

\end{abstract}

%
%

\ccsdesc[300]{Computing methodologies~Supervised learning by classification}
\ccsdesc[100]{Computing methodologies~Learning latent representations}
\ccsdesc[100]{Information systems~Query representation}


\keywords{Short text classification, memory networks, query expansion}

\maketitle

\section{Introduction}



Short texts make our lives easier as everyday Internet users. Search queries, short messages, microblogs, news headlines, and user comments efficiently convey, deliver, and disseminate information on various platforms. 
Short texts make our lives harder as data miners. Effectively making sense of short texts is critical for building many applications, such as search engines, recommender systems, social media services, and conversational agents. But it has always been challenging, due to the sparsity and ambiguity of information in short texts.

How does a human understand short texts? 
Consider the scenario where we read a headline ``\emph{Sequestration in Fiscal 2017}'', or encounter a paper titled ``\emph{Recognizing groceries {in situ} using {in vitro} training data}''~\cite{merler2007recognizing}, or see a Tweet ``\emph{Watching that movie makes me ROFL!}''. Without any context, they don't make clear sense for us either. Our spontaneous response is to open up a search engine and put the short text into the search box.
We believe that among billions of indexed Web pages, there exist relevant pages that elaborate the unfamiliar concepts in a short text. Even if the results are not all relevant, we know where to pay our attention and glean just the information we want. To achieve a thorough understanding, we sometimes update the query and perform another round of search.

The practice of a Web user querying on search engines is essentially leveraging a large amount of Web pages to understand short texts. Indeed, in literature, leveraging vast amounts of external data is proven to be an effective strategy for many applications of short text understanding such as query expansion \cite{zhai2001model, efron2012improving}, semantic relatedness analysis \cite{sahami2006web, gabrilovich2007computing}, short text classification \cite{phan2008learning, hu2009exploiting}, and question answering \cite{dumais2002web, schlaefer2011statistical}. In these applications, the features of short texts are typically expanded by selecting the most relevant documents from the entire corpus and then used in downstream tasks. A variety of heuristics are designed for the expansion process, which may or may not be optimal for downstream tasks.

Alternatively, we look for a principled process for short text expansion with a large collection of documents. Ideally, it would be able to emulate the human's information seeking process in search engines. Similar to the search engines, it should have a very efficient process to retrieve a list of documents that may be relevant to the short text. Instead of putting equal trust to all returned results, it should have a mechanism similar to humans' cognitive process, which can selectively pay attention to relevant results. It would be ideal to support iterative expansion of the short texts as a Web user may reformulate the query and conduct multiple rounds of search.




In this paper, we design such an automated process with a deep memory network, called the \emph{ExpaNet}. The network simulates the process of expanding a short text through \textit{searching} for relevant long documents, and is trained in an end-to-end fashion to optimize the downstream application. 
Given a short text, it first retrieves a set of potentially relevant documents from a large data collection, which may be noisy. Then, attention mechanisms~\cite{graves2014neural} are used to determine which documents are worth a closer look. Both soft and hard attention are considered, which either defines a probability distribution over the documents or focuses on an individual document. With the attention mechanism, new information from the long documents are identified, gathered, and integrated to reformulate the short text. In classical methods of query expansion, this is achieved by a linear combination of the two sources of information, and a global scalar coefficient is chosen for the combination based on heuristics or extensive tuning. Our network uses the Gated Recurrent Unit (GRU)~\cite{cho2014learning} as a principled way to combine the two sources of information, in which the weights are automatically determined for each short text. Similar to a human user who may continually update a search query for multiple rounds, our deep memory network also allows to expand the short text multiple times by using the reformulated short text as the next input. The final representation of short text is fed to a downstream learning task. In this paper, we take the task of short text classification as a demonstrative example. By optimizing the classification objective, the whole network is trained end-to-end by backpropagation.

We evaluate the proposed deep memory network using short texts of different genres, including titles of Wikipedia articles (general domain), titles of computer science publications (scientific domain), and tweets (social media domain). Experimental results show that it significantly outperforms classical text expansion methods and text classification methods that only use short text features.

To summarize, we make the following contributions:
\begin{itemize}
	\item We propose a novel end-to-end solution for short text expansion. A deep memory network-based model is designed to gather useful information from relevant documents and integrate it with the original short text. 
    \item We conduct extensive experiments on real-world short text data sets. Experimental results on short text classification show that our proposed deep memory networks significantly outperforms the classical query expansion methods and the methods which only use the features in short text.
\end{itemize}

\noindent \textbf{Organization.} The rest of this paper is organized as follows. Section 2 discusses the related work. Section 3 describes our end-to-end solution for short text expansion. Section 4 reports the experimental results, and we conclude the paper in Section 5. 

\section{Related Work}
Our work is related to three lines of research in literature: text representation, memory networks, and query/short text expansion.

\subsection{Text Representation}
Distributed representations of text have been proven to be very effective in various natural language processing tasks. These approaches can be roughly classified into two categories: unsupervised approaches (e.g., Skip-gram~\cite{mikolov2013efficient} and ParagraphVEC~\cite{le2014distributed}) and supervised approaches (e.g., convolutional neural networks (CNN)~\cite{kim2014convolutional}, recurrent neural networks (RNN)~\cite{graves2012supervised}, PTE~\cite{tang2015pte}, and FastText~\cite{joulin2016bag}). The representations learned by the unsupervised approaches are very general and can be applied to different tasks. However, their performance usually falls short on specific tasks since no supervision is leveraged when learning the representations. The supervised approaches have shown very promising results on different types of text corpus. For example, FastText~\cite{joulin2016bag} and PTE~\cite{tang2015pte} achieve state-of-the-art results for text classification on long documents while on short documents, CNN~\cite{kim2014convolutional} and RNN~\cite{graves2012supervised} achieve state-of-the-art results. All these approaches focus on learning text representations with the raw features, and no additional data is leveraged.

\subsection{Memory Networks}
Another line of related work are memory networks~\cite{sukhbaatar2015end,munkhdalai2016neural,kumar2015ask,graves2014neural}, which use the attention and memory mechanism in deep learning models. For example, Sukhbaatar et al. proposed an end-to-end memory network~\cite{sukhbaatar2015end} for question answering tasks. Given a question and contexts relevant to the question, the memory network employs a recurrent attention model over the contexts to iteratively identify relevant contexts for answering the question. Different variants of memory networks~\cite{kumar2015ask, munkhdalai2016neural} are proposed with different attention and memory updating mechanisms. 

Compared to these works, the current paper differs in several aspects: (1) most of the work on memory network focuses on question answering while our work studies a very different application: short text expansion and classification; (2) in the setting of question answering, the number of contexts for each question is very limited while in our setting, for each short text, the entire collection of documents are used as potential relevant contexts. (3) the soft attention mechanism is usually used in these works while in our work we investigate  both soft and hard attention mechanisms. In the experiments, we adapt the end-to-end memory networks to our task and compare it with our approach. 

\subsection{Short Text Expansion}
Our approach uses the search results from a large collection to expand short texts. This general strategy has been applied in many data mining tasks, notably query expansion with relevance feedback \cite{buckley1995automatic, zhai2001model, efron2012improving}, semantic relatedness analysis \cite{sahami2006web, gabrilovich2007computing}, short text classification \cite{phan2008learning, hu2009exploiting},  and question answering \cite{dumais2002web, schlaefer2011statistical}. 
The expanded text usually takes the form of interpolation between the original short text and the retrieved documents, which is then used in downstream tasks, such as retrieval and classification. Because the retrieved documents often contain noise, and the interpolation weights are often set by heuristics, the errors may accumulate in the pipeline and harm the performance of an end task.
This problem is known as query drift \cite{macdonald2007expertise} in query expansion.

Compared to previous work on short text expansion, we take an end-to-end approach to train the text expansion algorithm towards a clear learning objective. This turns short text expansion into an optimization problem and eliminates the need for extensive tuning of the interpolation weights \cite{lundquist1997improving}.

\section{Deep Memory Network for Short Text Expansion}
\begin{figure*}[htb!]
	\centering	\includegraphics[width=0.7\textwidth]{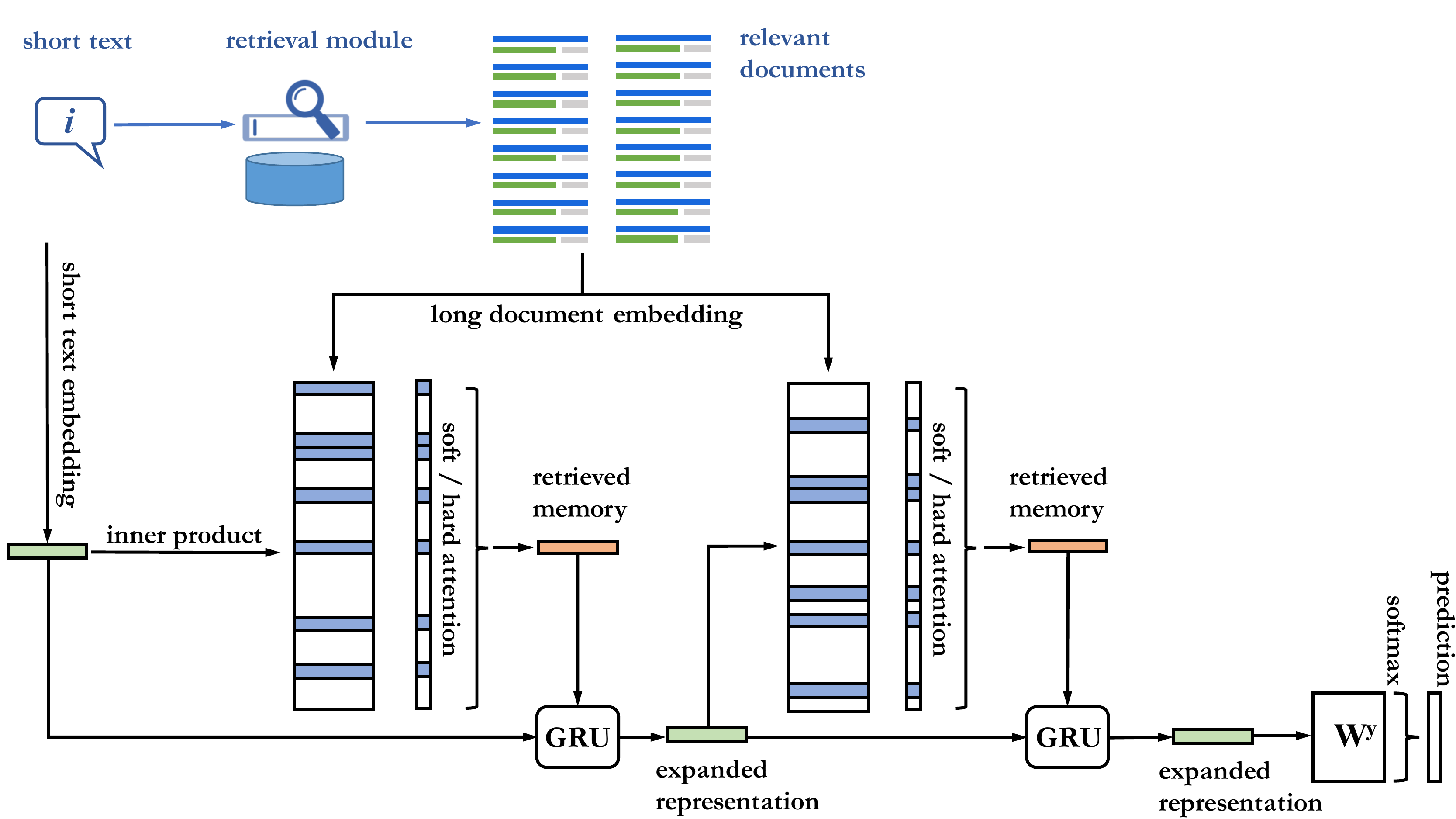}
	\caption{ExpaNet model structure  (2 hops). }
	\label{fig::model}
\end{figure*}
To understand a piece of short text, the common practice of a Web user is to formulate the short text as a search query, and then seek for definition, examples, paraphrases, and contexts in the returned Web pages. In other words, the Web user is leveraging the huge collection of Web pages for short text understanding. Therefore, in this paper, we study how to leverage external documents for expanding the representations of short texts and understanding its meaning. We take short text classification as an example of such task. Our problem is formally defined as follows:

\begin{definition}
	\label{def::problem_definition}
	\textbf{(Problem Definition.)}
	\textsl{Given a collection of long documents $\calC$, we aim to learn a function $f$ that expands a short text $q$ into a richer representation $q'$, i.e., $q'=f(q, \calC)$. Based on the richer representation $q'$, we can accurately classify the short text into one of the predefined categories $\calY$.
}
\end{definition}

Leveraging long documents for short text expansion has been widely studied in  information retrieval literature, where a query is expanded by leveraging the top-K documents returned by the retrieval systems, a method known as pseudo relevance feedback~\cite{zhai2001model}.  In such a method, top-K documents are usually retrieved with a search function, and then some terms are selected from the top-K documents using heuristic methods such as TFIDF weighting and probability weighting, which are added back to the original query. However, these methods are not accurate since the returned top-K documents and terms selected from the top-K documents can be noisy and cause topic drift. 

In this section we introduce ExpaNet, an end-to-end solution based on deep memory network. Our approach shares similar intuition with pseudo-relevance feedback but is much more principled. It can be trained to automatically identify the relevant documents for the given query (short text) and filter out non-relevant ones. 

ExpaNet  integrates five different modules including retrieval module, short text representation module, long document representation module, expansion module, and classification module. Since the long document collection $\calC$ can be huge, the \emph{retrieval module} provides an efficient way to retrieve a small subset of potentially relevant documents $\calC_q$ for the given short text $q$. The \emph{short text representation module} maps the short text $q$ into a continue representation $\vec{q}$. The \emph{long document representation module} represents each document $d \in \calC_q$ with a continuous representation $\vec{d}$ and put it into the memory $M$. The \emph{expansion module} expands the $\vec{q}$ into a new representation $\vec{q}'$ by leveraging the memory $M$ using multiple hops. Finally, the \emph{classification module} predicts the category with the expanded representation $\vec{q}'$ as input. All these modules are coupled together and trained through error backpropagation. Next, we introduce these different modules respectively.

\subsection{{Retrieval Module}}
In practice, the external long document collections $\calC$ can be very large. For example, the entire Web contains billions of Web pages, and the entire Wikipedia contains millions of articles. For a short text $q$, only a few documents from the collection $\calC$ would be relevant to the query. Therefore, we first use the original short text $q$ as a query to search for a set of potentially relevant long documents $\calC_q$ from an external large collection $\calC$. The documents will be used by the model as the ``raw material'' for text expansion. The goal of this step is to obtain relevant documents efficiently and with high recall. This process can be implemented efficiently with existing techniques such as an inverted index used in information retrieval, locality sensitive hashing for high-dimensional data points, and directly making use of APIs provided by existing search engines. To ensure a high recall, one can set the number of returned documents to be reasonably large, e.g., tens or hundreds of documents.

\subsection{{Short Text Representation Module}}
We represent each short text $q=w_1,\ldots, w_n$ as a $d$-dimensional vector $\vec{q}$ in a continuous space. Each word in the vocabulary is represented as a $d$-dimensional vector, and then the entire short text is represented as the average vector of words in the short text, i.e.,
\begin{align}
\vec{q} = \frac{\sum_{i=1}^n \bA_{w_n}}{n}
\end{align}
where $\bA \in \bbR^{d \times V}$ is the word embedding matrix, $V$ is the size of the vocabulary. There could be more sophisticated ways to encode a piece of text (such as with convolutional neural networks~\cite{kim2014convolutional} or recurrent neural networks~\cite{palangi2016deep}). We choose the simple averaging approach as it was shown to work well in our previous work \cite{tang2015pte} and much easier to train.

\subsection{{Long Document Representation Module}}
Each long document is also represented as a $d$-dimensional vector space. Similarly, each document $d_i=w_1,\ldots,w_n$ are represented as the average vector of the words in the documents, i.e., 

\begin{align}
\vec{d_i} = \frac{\sum_{i=1}^n \bB_{w_n}}{n},
\end{align}
where $\bB \in \bbR^{d \times V}$ is word embedding matrix for long document representations. 

\subsection{{Expansion Module}}
The expansion module is the core part of ExpaNet. The goal is to expand the continuous representation of input short text $\vec{q}$ by incorporating the information in the memory $M=\{\vec{d}_i\}_{i=1}^K$, where $K$ is the number of documents in the memory. The expansion process can be divided into two different components: (1) given the query representation $\vec{q}$, what information should we read from the memory? (2) how to integrate the information from the memory with the original query representation $\vec{q}$?

\subsubsection{Memory Reading}
For memory reading component, we aim to identify the relevant documents to the given query $\vec{q}$. Here, we use the attention mechanism. Two types of attention mechanisms are used: soft attention~\cite{sukhbaatar2015end} and hard attention~\cite{mnih2014recurrent}.

\textbf{Soft attention}: Soft attention is widely used in existing memory networks. We use the same mechanism as~\cite{sukhbaatar2015end}. The relevance between the query $\vec{q}$ and each document $\vec{d}_i$ is calculated as their inner product, and a softmax function is used to define the attention probability over each document $i$ in the memory, i.e., 

\begin{align}
a_i = \text{Softmax}\paren{ \vec{q}^{\top} \vec{d}_i } \ ,
\end{align}
where $\text{Softmax}(z_i) = e^{z_i}/\sum_j e^{z_j}$. In this way, the $a_i$'s define a probability distribution over the long documents in memory $M$, and the information read from the document is defined as:
\begin{align}
\vec{o}=\sum_{i=1}^K a_i\vec{d}_i \ .
\end{align}

\textbf{Hard attention}: 
Instead of looking at each document with some probability, a human searcher often picks a document that seems relevant and focus on it. Therefore, we also investigate using hard attention here~\cite{mnih2014recurrent}, which is achieved by randomly sampling a document from the probability distribution $\vec{a}=(a_1, \ldots, a_K)$ defined in the soft attention, i.e., 
\begin{align}
 \vec{p} \sim \text{multinomial}(\vec{a}) \ ,
\end{align}
where $\vec{p}$ is a one-hot vector. Then the information read from the memory is defined as:
\begin{align}
\vec{o}=\sum_{i=1}^K p_i\vec{d}_i.
\end{align}

However, as mentioned in~\cite{mnih2014recurrent}, training a hard attention model is very hard, which has a high variance of the gradients (e.g., the REINFORCE~\cite{williams1992simple} algorithm), and complicated variance reduction methods~\cite{xu2015show} must be used. In this paper, we use a recent technique, the Gumbel-Softmax \cite{jang2016categorical}, for backpropagating through samples, which has a low gradient variance. Specifically, each sample is drawn according to the following distribution:

\begin{align}
p_i = \text{Softmax}\paren{ \frac{ \vec{q}^{\top} \vec{d}_i + g_i}{\tau} } \ ,
\end{align}
where $g_i$ follows the {Gumbel}(0,1) distribution, and $\tau$ is the temperature hyperparameter ($\tau$ is set as 2.0 in our experiments). For more details about the Gumbel-softmax distribution, readers can refer to~\cite{jang2016categorical}.

\subsubsection{Short Text Expansion.} With the \emph{memory reading} component, the model is able to retrieve relevant information, i.e., $\vec{o}$, from the memory. Then how should we reformulate the short text? It is natural to integrate information from the  original representation $\vec{q}$,  and information retrieved from the memory $\vec{o}$. Indeed, a typical method for query expansion in information retrieval is to interpolate between the original query and the expanded document \cite{rocchio1971relevance,zhai2001model}, where a scalar parameter is used to trade off between the two information and empirically tuned, which is based on heuristics. Here, we use a principled method to integrate the two sources of information. We use a gating mechanism, the Gated Recurrent Unit (GRU)~\cite{cho2014learning}, to combine the information, which is able to automatically determine the weight of the two sources of information. Specifically, the two sources of information are integrated as follows:
\begin{align}
\vec{z} & = \sigma\paren{ \bW^{(z)} \vec{q} + \bU^{(z)} \vec{o} } \ ; \\
\vec{r} & = \sigma\paren{ \bW^{(r)} \vec{q} + \bU^{(r)} \vec{o} } \ ; \\
\vec{o}' & = \tanh\paren{\bW \vec{q} + \vec{r} \circ \bU \vec{o} } \ ; \\
\vec{q}' & = (\mathbf{1} - \vec{z}) \circ \vec{q} + \vec{z} \circ \vec{o}' \ ,
\end{align}
where $\circ$ denotes elementwise multiplication and $\sigma(x) = 1/(1+\exp(x))$, $\tanh(x) = (1-\exp(-2x))/(1+\exp(-2x))$ are both elementwise operations. $\vec{o}'$ is the new information from the memory, which is determined by both sources of information $\vec{q}$ and $\vec{o}$. $\vec{z}$ is the weighting vector between the original information $\vec{q}$ and the new information $\vec{o}'$. The output $\vec{q}'$ is the expanded representation of the input short text $q$. 

\subsubsection{Iterative Expansion via Multiple-hops.} When a Web user 
inputs a query and reads the search results, the user may reformulate the query. This process can continue several times until the user understands the query. Our algorithm also tries to simulate this process, which is achieved through the recurrent attention mechanism using multiple hops in the \emph{short text expansion} component. More specifically, when a expanded representation $\vec{q}'$ is output by the \emph{short text expansion} component, the representation $\vec{q}'$ is treated as an initial query to the module. This process can be repeated several times, and the final output representation is used as the representation of the original query $q$.

In practice, when the query is updated, one may ask that the set of relevant documents should be re-retrieved from the entire collections. However, as mentioned previously, the initial set of retrieved documents have a very high recall, which means that the relevant documents to the new query are very likely to belong the initial retrieved set. Only the weights between the query and the documents need to be updated, which is taken care of by the attention mechanism. 

\subsection{Classification Module} As in classical methods for query expansion, we keep the original short text representation $\vec{q}$ and represent the final short text representation as a concatenation of $\vec{q}$ and the expanded representation $\vec{q}'$, i.e., $\vec{q}_{\text{final}}=[\vec{q}, \vec{q}']$, which is then used to predict the category of the short text. A fully connected layer is first applied to the short text representation and then followed by a Softmax transformation, yielding a distribution over the categories, i.e., 
\begin{align}
p(y|\vec{q}_{\text{final}})=\text{Softmax}(\bW^y \ \vec{q}_{\text{final}}),
\end{align}
where $\bW^y \in \bbR^{\abs{\calY} \times 2d}$ is the parameter for fully connected layer. 

\subsection{Training}
In this paper, we take the example of short text classification as the goal of short text expansion. Therefore, the ultimate goal is to accurately predict the category of the short text, and the cross entropy loss function is used. Specifically, given a training data set $(q_i, y_i)$ and a document collection $\calC$,  we aim to minimize the loss:
\begin{align}
\sum_{i=1}^n \sum_{y\in \calY} \ind{y=y_i} \log p\paren{y|q_i, \calC} ,
\end{align}
where $p\paren{y|q_i, \calC}$ is the probability of class $y$ given short text $q_i$ and long document collection $\calC$, predicted by the network. The whole network is trained by backpropagation including the word embeddings $\bA$, $\bB$, weights of fully connected network $\bW^y$ for classification, and parameters $\bW^{(z)}, \bU^{(z)}, \bW^{(r)}, \bU^{(r)}, \bW, \bU$ in the GRU.

\section{Experiments}
In the experiments, we compare our algorithm (ExpaNet) to state-of-the-art methods for short text classification and classical methods of query expansion. On three real world data sets, ExpaNet shows superior performance. We also analyze the effect of retrieval collection choice, parameter sensitivity, and attention distribution.

\subsection{Data Sets}
\label{sec:dataset}
We test our algorithm on three different genres of short texts. 
Basic statistics of these data sets are summarized in Table \ref{tab:short_doc_stats} and \ref{tab:long_doc_stats}.

{\sc Wikipedia}. Titles of Wikipedia articles represent short texts in the general domain. The length of Wikipedia titles is on average 3.12 words, similar to that of search queries \cite{bendersky2009analysis}. We take a recent snapshot of English Wikipedia\footnote{\url{https://dumps.wikimedia.org/enwiki/20161120}} to construct this data set. We use 15 categories in the main topic classifications of Wikipedia\footnote{\url{https://en.wikipedia.org/wiki/Category:Main\_topic\_classifications}} as our labels: ``Arts'', ``Games'', ``Geography'', ``Health'', ``History'', ``Industry'', ``Law'', ``Life'', ``Mathematics'', ``Matter'', ``Nature'', ``People'', ``Religion'', ``Science and Technology'', and ``Society''. 
We assign a title to its closest category in terms of geodesic distance (shortest path length) in the graph of Wikipedia categories. To generate multiclass classification data set, we include only titles with a unique category, i.e. only that category has the shortest geodesic distance to the article. To construct a collection of semantically related long documents, we use the the abstract of all Wikipedia articles.

{\sc Dblp}. Titles of computer science literature represent short texts in formal communication. We choose 6 diverse research fields for classification, including ``Database'', ``Artificial Intelligence'', ``Hardware'', ``Systems'', ``Programming Languages'', and ``Theory''. For each field, we select representative conferences and collect the titles of papers published in these conferences as labeled data. To construct a collection of semantically related long documents, we use the abstracts of all papers in DBLP bibliography database. \footnote{\url{http://aminer.org/lab-datasets/citation/DBLP-citation-Jan8.tar.bz2}}

{\sc Twitter}. The 140-character microblog data represent informal short texts widely used in social media. We use a large corpus of tweets for positive/negative sentiment classification.\footnote{\url{http://thinknook.com/twitter-sentiment-analysis-training-corpus-dataset-2012-09-22/}} We randomly sampled 1,200,000 tweets and split them into training and test sets.
Because tweets is special genre of text, it is non-trivial to obtain semantically related long documents. Since the data set itself is reasonably large, we use the training set as the document collection.


We use the Apache Lucene library \footnote{\url{https://lucene.apache.org}} to construct full-text index for each document collection. This allows efficient document retrieval using short texts as queries. For each short text, we associate top $K$ relevant documents as its \emph{memory}. We use Dirichlet smoothing language modeling  as the retrieval function \cite{zhai2001study}.

\begin{table}[htbp]
  \caption{Statistics of short text data sets}
  \label{tab:short_doc_stats}
  \begin{tabular}{cccc}
    \toprule
     & {\sc Wikipedia} & {\sc Dblp} & {\sc Twitter}\\
    \midrule
    Train         & 18,000  & 61,479 & 800,000 \\
    Test          & 12,000  & 20,000 & 400,000 \\
    Vocabulary         & 25,550  & 22,686 & 535,997\\
    Avg. doc. length   & 3.12       & 9.48      & 13.69      \\
    \# of classes & 15      & 6      & 2      \\
  \bottomrule
\end{tabular}
\end{table}

\begin{table}[htbp]
  \caption{Statistics of long document collections}
  \label{tab:long_doc_stats}
  \begin{tabular}{cccc}
    \toprule
     & {\sc Wikipedia} abstract & {\sc Dblp} abstract \\
    \midrule
    \# of docs    & 4,747,988  & 480,558  \\
    Vocabulary         & 3,768,403 & 310,178  \\
    Avg. doc. length   & 98.37  & 138.68      \\
  \bottomrule
\end{tabular}
\end{table}

We use standard classification performance metrics to evaluate  different algorithms: micro-averaged F1 and macro-averaged F1.


\subsection{Experimental Setup}


\subsubsection{Compared Methods.} We compare three different types of methods: (1) existing typical approaches for text representations, which only use the information in the original short text; (2) classical relevance feedback methods for query expansion in information retrieval, which leverages external documents to improve short text representation; (3) end-to-end short text expansion solution based on memory networks, which also leverage external documents and are trained in an end-to-end fashion.  

\textbf{Text representation approaches} can be grouped as follows.

\emph{Unsupervised methods}: (1) The classical ``bag-of-words'' represenation (\textbf{BOW}). Each document is a $|V|$-dimensional vector, where each dimension is the TFIDF representation a word. (2)  \textbf{Skip-gram}: the state-of-the-art word embedding model \cite{mikolov2013distributed}; (3) \textbf{LINE}: the large-scale information network embedding model \cite{tang2015line}. It is used to learn unsupervised word embeddings from word co-occurrence networks and word-document networks. We take average of word embeddings to produce a document embedding.

\emph{Supervised methods}: (1) \textbf{PTE}: the predictive text embedding model \cite{tang2015pte}. We use the PTE model to learn supervised word embeddings from word co-occurrence networks, word-document networks, and word-label networks. We take average of word embeddings to produce a document embedding.  (2) \textbf{CNN}: the supervised text embedding model based on convolutional neural networks \cite{kim2014convolutional}. (3) \textbf{RNN}: the supervised text embedding model based on recurrent neural networks with bidirectional GRUs. (4) \textbf{FastText}: a supervised text embedding model showing comparable performance to more sophisticated deep learning models \cite{joulin2016bag}.

\textbf{Query expansion methods}: For classical query expansion algorithms, we consider Rocchio's method \cite{rocchio1971relevance}. Both short text and long documents are represented as sparse ``bag-of-words'' vectors with TFIDF weighting. We treat the long documents $\vec{d_i}$ as pseudo relevant documents, and expand the short text $\vec{q}$ by interpolating between $\vec{q}$ and the average of $\vec{d_i}$'s:
\begin{align}
\vec{q'} = (1 - \lambda) \ \vec{q} +  \frac{\lambda}{K} \sum_{i=1}^K \vec{d_i} \ , \label{equ:rocchio}
\end{align}
where $\lambda \in [0,1]$ is the interpolation weight. $\lambda$ is tuned for each data set on a validation set. We call this method \textbf{BOW}$_{{RF}}$. 

Another straightforward query expansion approach is to concatenate the short text and its long documents together, and use existing methods to learn text representation for this pseudo-document. Specifically,
we use Skip-gram, LINE, and PTE to learn word embeddings, and generate the pseudo-document vector by averaging word embeddings. This gives us three variants: \textbf{Skip-gram}$_{{RF}}$, \textbf{LINE}$_{{RF}}$, and \textbf{PTE}$_{{RF}}$, corresponding to three text representation methods.

\textbf{Memory network-based methods}: we adapt the original memory network  \cite{sukhbaatar2015end}  to our problem setting (\textbf{MemNet}). We treat the short text as the question, the documents retrieved by our retrieval module as memories, and the target category as the answer. Finally, we include two variants of our algorithm: the short text expansion memory network with soft attention (\textbf{ExpaNet-S}) and hard attention (\textbf{ExpaNet-H}).

We consider two settings of retrieved documents: (1) \textbf{short text memory}: for each short text in training and test set, the memories are short texts pre-retrieved from the training set. (2) \textbf{long document memory}: for each short text in training and test set, the memories are long documents pre-retrieved from an external document collection. As mentioned in Section \ref{sec:dataset}, we only consider short text memory for the {\sc Twitter} data set.

\subsubsection{Parameter Settings}
To prepare relevant documents for each short text, we retrieve top 20 results from the document collection returned by Dirichlet smoothing language modeling ($\mu = 2000$, default in Lucene). Considering  Web search engines typically show 10 results per page,  20 results is reasonably large for a high recall of relevant documents. In rare cases where the retrieval function does not return enough results, we randomly duplicate the returned results to make 20 documents. 

To train text classification models, 
we use one-vs-rest multiclass support vector machines with linear kernel implemented in the LibLinear package \cite{fan2008liblinear} with regularization weight $c = 1$. For text embedding methods, we set embedding dimension to be 100. To represent a piece of text as a sequence of words, we use the first 15 words for short text and the first 100 words for long documents; zero-padding is used when the text is not long enough. End-to-end learning algorithms are trained using Adam stochastic gradient descent algorithm \cite{kingma2014adam}, in mini-batches of size 32. The initial learning rate is empirically set to be $10^{-3}$ for CNN and RNN, and $10^{-2}$ for FastText and memory network-based methods.
Word embedding matrices are initialized with numbers drawn from $\mathcal N(0,0.1^2)$.

For each of the three memory network-based methods, the number of hops is tuned on hold-out data sets ($\#hops \in \set{1,2,3,4}$).

\begin{table*}[htbp!]
  \caption{Classification performance of compared methods }
  \label{tab:overall_performance}
  \begin{tabular}{cccccccc}
    \toprule
 Settings & Methods & \multicolumn{2}{c}{{\sc Wikipedia}} & \multicolumn{2}{c}{{\sc Dblp}} & \multicolumn{2}{c}{{\sc Twitter}} \\
  &  &   Micro-F1 & Macro-F1   &    Micro-F1 & Macro-F1   &   Micro-F1 & Macro-F1 \\
  \midrule
            &  BOW         & 42.63 & 42.37   &   75.28 & 71.59   &   75.27 & 75.27  \\
            &  Skip-gram   & 26.75 & 25.90   &   73.08 & 68.92   &   73.02 & 73.00 \\
 short text &  LINE        & 34.93 & 32.84   &   73.98 & 69.02   &   73.19 & 73.18 \\
 only       &  PTE         & 38.78 & 38.58   &   76.45 & 72.74   &   73.80 & 73.80  \\
            &  CNN         & 41.70 & 41.68   &   77.38 & 74.35   &   77.83 & 77.23 \\
            &  RNN         & 42.40 & 42.10   &   77.84 & 74.83   &   77.45 & 76.72 \\
            &  FastText    & 43.33 & 42.79   &   77.15 & 74.05   &   74.20 & 74.07 \\
  \midrule
            &  BOW$_{RF}$         & 43.01 & 42.89   &   77.45 & 74.16   &   76.21 & 76.19  \\
            &  Skip-gram$_{RF}$   & 34.57 & 32.84   &   75.68 & 71.84   &   73.14 & 73.11  \\
 short text &  LINE$_{RF}$        & 35.05 & 33.19   &   75.92 & 72.14   &   73.20 & 73.20  \\
 memory     &  PTE$_{RF}$         & 39.54 & 39.17   &   77.37 & 73.84   &   73.93 & 73.91  \\
            &  MemNet     & 42.20 & 41.74   &   77.94 & 74.75   &   77.76 & 77.44    \\
            &  ExpaNet-S    & 43.67$^{***}$ & 43.60$^{***}$   &   79.05$^{***}$ & 76.09$^{***}$   &   78.55 & 78.24   \\
            &  ExpaNet-H    & 42.37 & 42.31$^{**}$   &   78.94$^{***}$ & 75.97$^{***}$   &   79.25$^{***}$ & 78.91$^{***}$   \\
  \midrule
            &  BOW$_{RF}$         & 47.13 & 47.12   &   78.26 & 75.25   & - & - \\
            &  Skip-gram$_{RF}$   & 46.66 & 45.54   &   75.55 & 71.93   & - & - \\
 long doc.  &  LINE$_{RF}$        & 46.52 & 45.36   &   75.75 & 72.19   & - & - \\
 memory     &  PTE$_{RF}$         & 48.15 & 47.43   &   78.31 & 75.26   & - & - \\
            &  MemNet     & 47.66 & 47.57   &   79.16 & 75.91   & - & - \\
            &  ExpaNet-S    & 50.85$^{***}$ & 50.69$^{***}$   &   80.32$^{***}$ & 77.60$^{***}$   & - & - \\
            &  ExpaNet-H    & 50.68$^{***}$ & 50.50$^{***}$   &   80.12$^{***}$ & 77.35$^{***}$   & - & - \\
  \bottomrule
\end{tabular}

$^{**}$($^{***}$) means the result is significant according to Student's T-test at level 0.05 (0.01) compared to MemNet. 
\end{table*}

\subsection{Overall Performance}

We compare three different types of approaches: methods with only short text features, short text expansion using classical query expansion methods, and end-to-end short text expansion methods based on memory networks. For the latter two types of methods, besides leveraging external long documents as memory, we also investigate the effect of treating the training short documents as memory. For all the methods, the performance are averaged over five runs with random parameter initialization. Table~\ref{tab:overall_performance} summarizes the performance of all compared methods. Statistical significance of the results are provided by comparing our methods to the original memory networks~\cite{sukhbaatar2015end}, a strong baseline method. In general, our methods significantly outperforms the original memory networks and baseline methods only using original features of short texts and classical query expansion methods.

For the methods with only short text features, the unsupervised approaches BOW, Skipgram,and LINE do not perform well since no supervision is used to learn the representations. For the supervised approaches, the PTE and FastText, which ignore the order of the words, perform comparably as CNN and RNN on topic classification tasks ({\sc Dblp} and {\sc Wikipedia}). However, the performance is significantly inferior to CNN and RNN on the task of sentiment classification on the Twitter data set, for which the order of the words is very important.

For query expansion methods, additional information from relevant documents tends to improve classification performance. The classical ``bag-of-words'' representation performs very well. Long document memory provide richer information than short document memory, and is more effective in general. The performance gain is most salient when the text is extremely short and the relevant documents are long ({\sc Wikipedia}).

Compared to the classical query expansion methods, the performance of memory network-based methods are significantly better since the expansion process of classical methods are heuristic, which may introduce noise into the original representation while the memory network based solution provide an end-to-end solution. Our model with either soft or hard attention mechanism outperforms the original memory networks since the GRU unit is used to integrate the  information from the original query and information read from memory. The performances of soft and hard attention are on par with each other. On two data sets ({\sc Wikipedia} and {\sc Dblp}), soft attention performs slightly better than hard attention. On {\sc Twitter}, hard attention performs slightly better than soft attention.

\subsection{Comparison of Expansion Using  General v.s. Specific Domain Long Documents}
We often have access to abundant long documents in general domain, such as Wikipedia and the World Wide Web, but less so for specific domains. These long documents may not exactly match the domain of a short text classification task, but some of them may still provide relevant information as long as there is some overlap in semantics and genre. To test this hypothesis, we take titles of {\sc Dblp} and search for relevant documents in {\sc Wikipedia} as the memory for classification. 
Intuitively, we are testing if ``reading computer science literature could be as useful as reading relevant articles in Wikipedia''.
The classification performance is shown in Table \ref{tab:dblp_searchWIKI}.

Overall, we observe a performance drop after using {\sc Wikipedia} abstracts instead of  {\sc Dblp} abstracts as memory. The performance drop is more salient when we use unsupervised text representation, because long documents in general domain necessarily introduce noise. Supervised representation learning methods can mitigate noise by fine-tuning text representation towards the task. Memory networks-based methods have the smallest performance gap by providing an end-to-end solution, among which our algorithm (ExpaNet) performs the best.

\begin{table*}[htbp!]
  \caption{Performance comparison of expansion using long documents from general domain vs. specific domain }
  \label{tab:dblp_searchWIKI}
  \begin{tabular}{ccccccc}
    \toprule
   Methods &  \multicolumn{2}{c}{{\sc Dblp} memory} & \multicolumn{2}{c}{ {\sc Wikipedia} memory } & \multicolumn{2}{c}{Performance gap} \\
    &       Micro-F1 & Macro-F1   &   Micro-F1 & Macro-F1   &   Micro-F1 & Macro-F1 \\
    \midrule
    BOW$_{RF}$            &   78.26 & 75.25   &   71.00 & 67.55    &   -7.26 & -7.70 \\
    Skip-gram$_{RF}$      &   75.55 & 71.93   &   68.63 & 64.73    &   -6.92 & -7.20 \\
    LINE$_{RF}$           &   75.75 & 72.19   &   68.95 & 65.15    &   -6.80 & -7.04 \\
    PTE$_{RF}$            &   78.31 & 75.26   &   75.78 & 72.35    &   -2.53 & -2.91 \\
    MemNet        &   79.16 & 75.91   &   77.18 & 74.04    &   -1.98 & -1.87 \\
    ExpaNet-S       &   80.32$^{***}$ & 77.60$^{***}$   &   78.34$^{***}$ & 75.52$^{***}$     & -1.98 & -2.08 \\
    ExpaNet-H       &   80.12$^{***}$ & 77.35$^{***}$   &   78.04$^{**}$ & 75.15$^{**}$       & -2.08 & -2.20 \\
  \bottomrule
\end{tabular}

$^{**}$($^{***}$) means the result is significant according to Student's T-test at level 0.05 (0.01) compared to MemNet.
\end{table*}
\subsection{Parameter Sensitivity}

\subsubsection{Effect of number of hops.}
\begin{figure*}[htb!]
\begin{minipage}[t]{1.0\textwidth}
\begin{minipage}[t]{0.32\textwidth}
\centering
\includegraphics[width=1\textwidth]{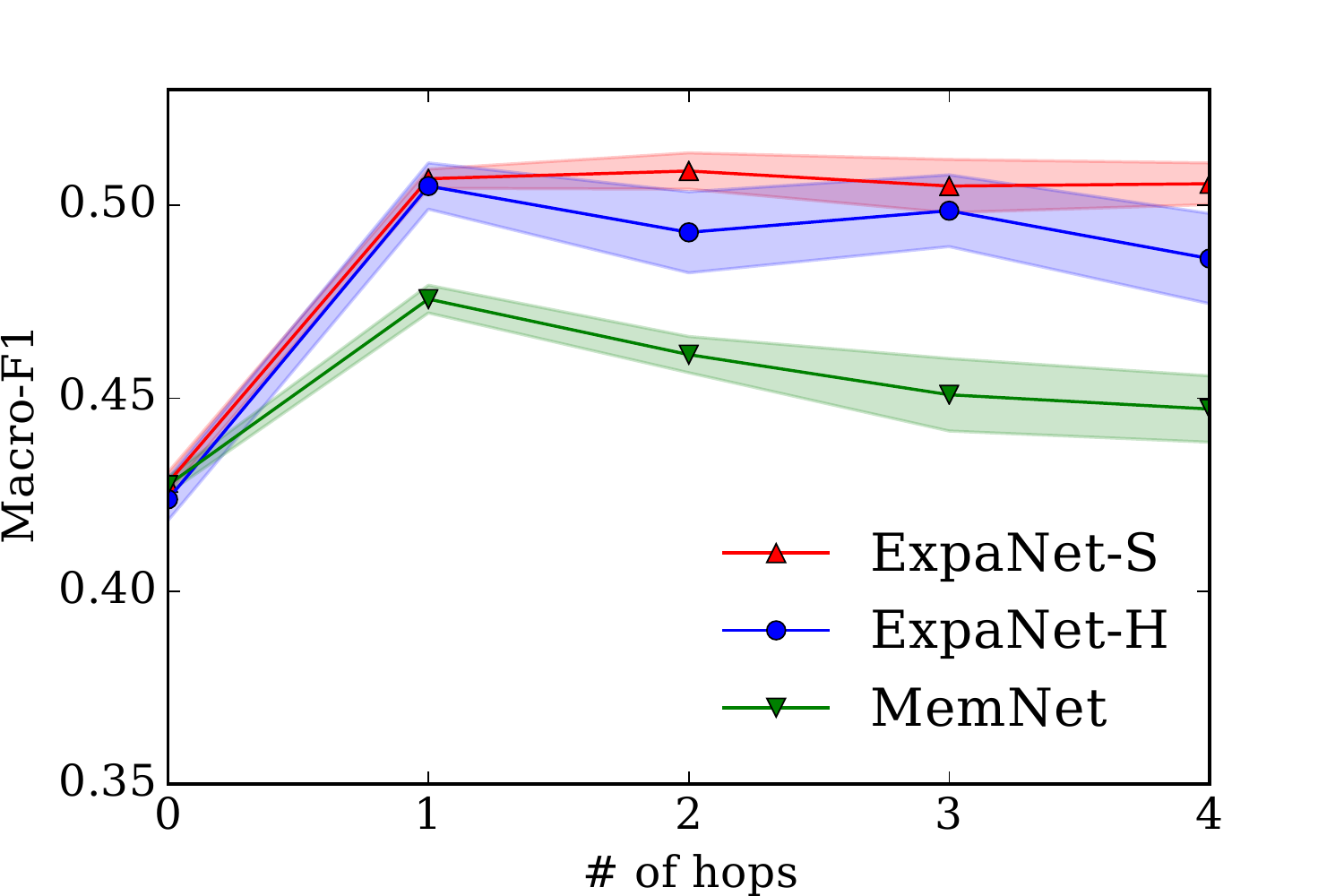}\\ 
(a) {\sc Wikipedia}
\end{minipage}
\begin{minipage}[t]{0.32\textwidth}
\centering
\includegraphics[width=1\textwidth]{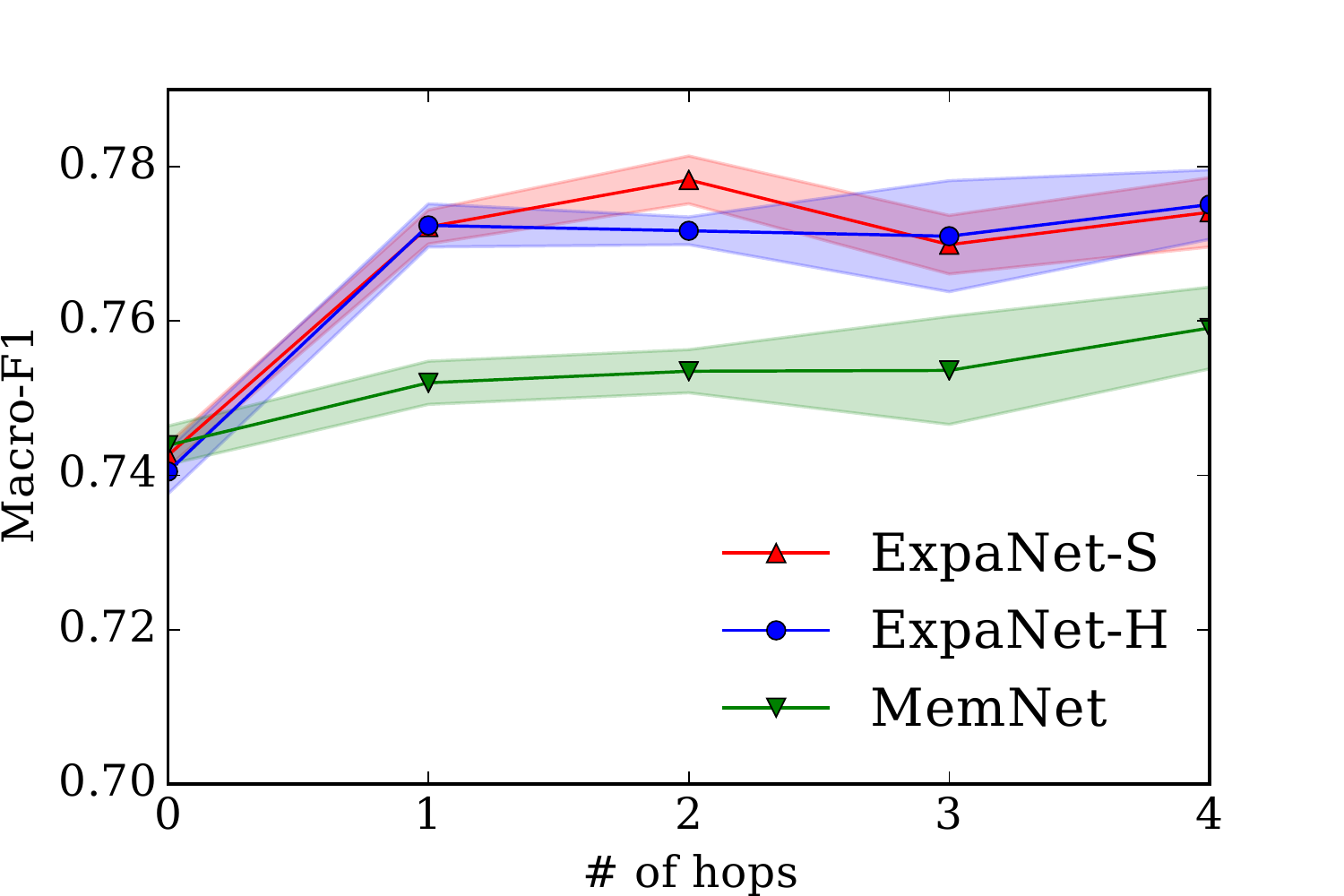}\\
(b) {\sc Dblp}
\end{minipage}
\begin{minipage}[t]{0.32\textwidth}
\centering
\includegraphics[width=1\textwidth]{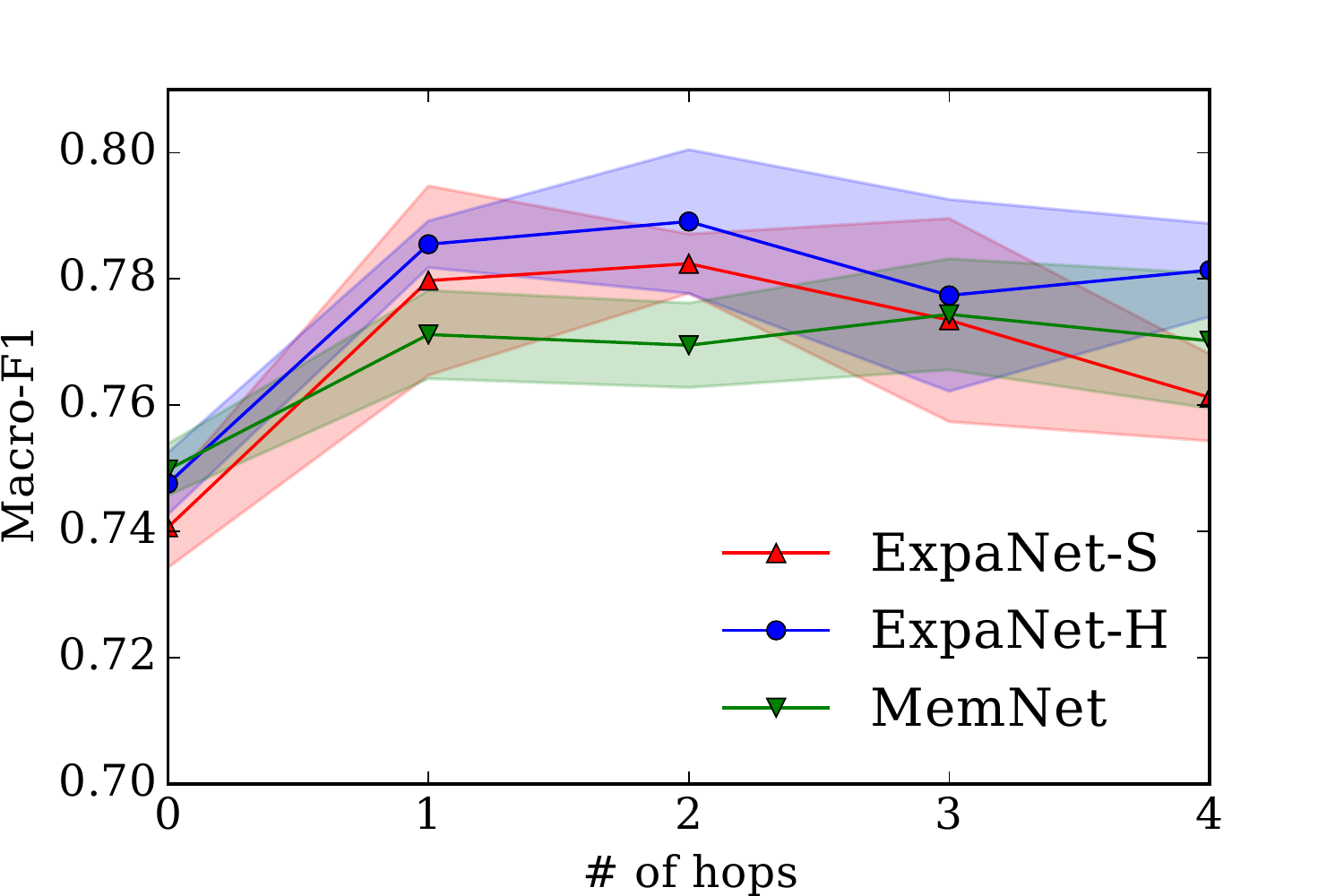}\\
(c) {\sc Twitter}
\end{minipage}
\caption{Performance w.r.t. \# of hops. When \#hops=0, MemNet, ExpaNet-S, and ExpaNet-H are the same model since none of them use the memory. Color regions correspond to $\pm 1$ standard deviation around the mean.}
\label{fig:num_hubs}
\end{minipage}
\end{figure*}

In our model, each hop selectively incorporates information from memory and refines the representation of short text from previous hop.  Naturally we raise the hypothesis that more hops will further refine short text representation and lead to increased performance. To test the hypothesis, we train the memory networks with different number of hops (0,1,2,3,4) and observe the macro-averaged F1 score on the three data sets. To obtain statistical confidence, we run each configuration 5 times and compute the mean and standard deviation. Figure \ref{fig:num_hubs} shows the performance curve as the number of hops increases. 


We see that as the number of hops increases, classification performance will increase and then quickly saturate. A large gain can be observed from \#hops=0 to 1, which introduces the most information, and a slight gain from \#hops=1 to 2. The performance becomes somewhat unstable with even more hops, which may be due to over-fitting. Ideally, the number of hops is different for different short text: some need more refinements as they are too short, while others need less. We leave how to \emph{learn} the best number of hops for expanding each short text as our future work.

\subsubsection{Effect of memory size.}
In the \emph{retrieval module}, we retrieve a set of a relevant documents, which are put into the memory. In this part, we study the effect of memory size with respect to classification performance. We take the ExpaNet-soft with 1 hop as an example and vary the number of memory cells $K$. All the results are averaged over five runs with random initialization.

\begin{figure*}[htb!]
\begin{minipage}[t]{1.0\textwidth}
\begin{minipage}[t]{0.32\textwidth}
\centering
\includegraphics[width=1\textwidth]{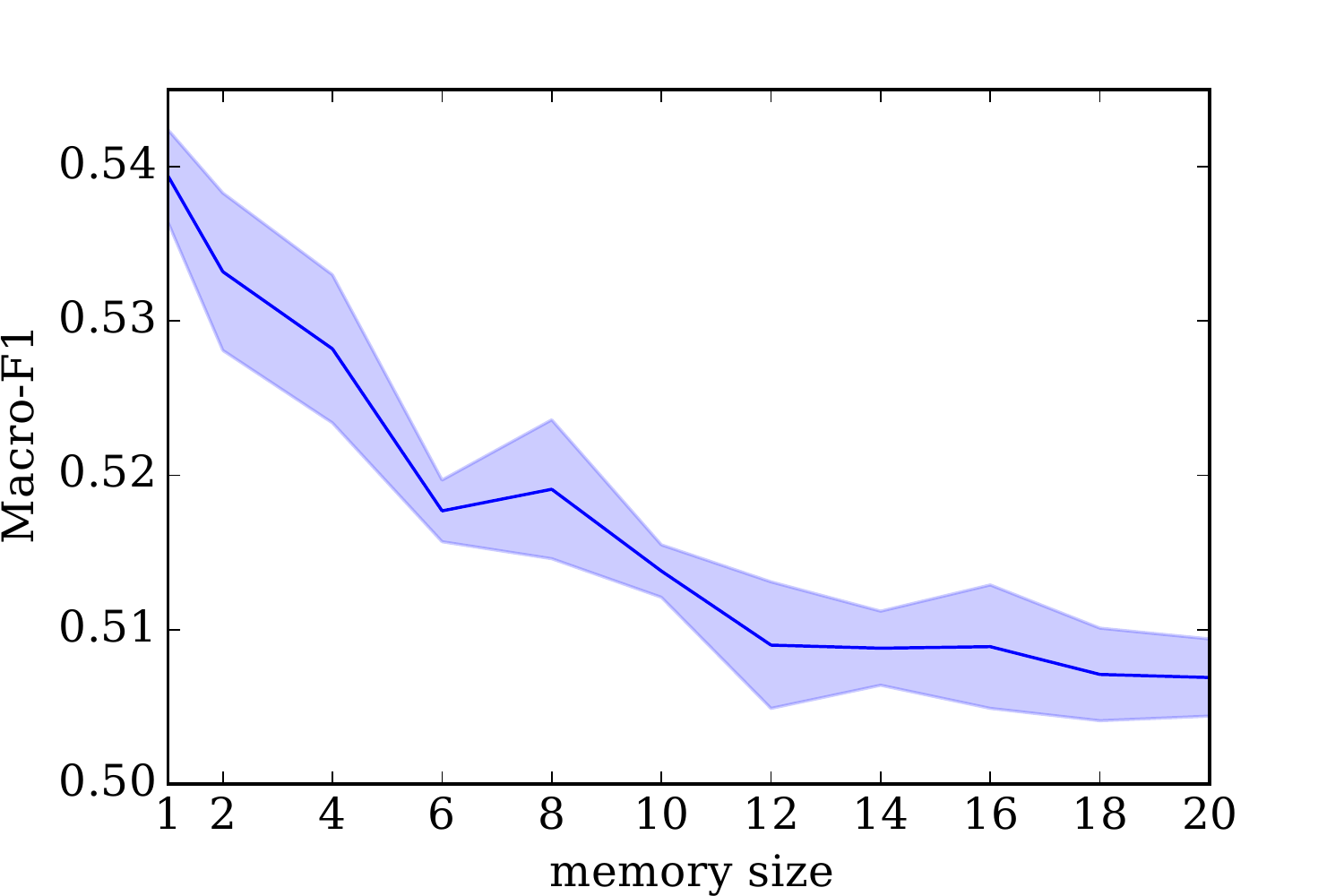}\\
(a) {\sc Wikipedia}
\end{minipage}
\begin{minipage}[t]{0.32\textwidth}
\centering
\includegraphics[width=1\textwidth]{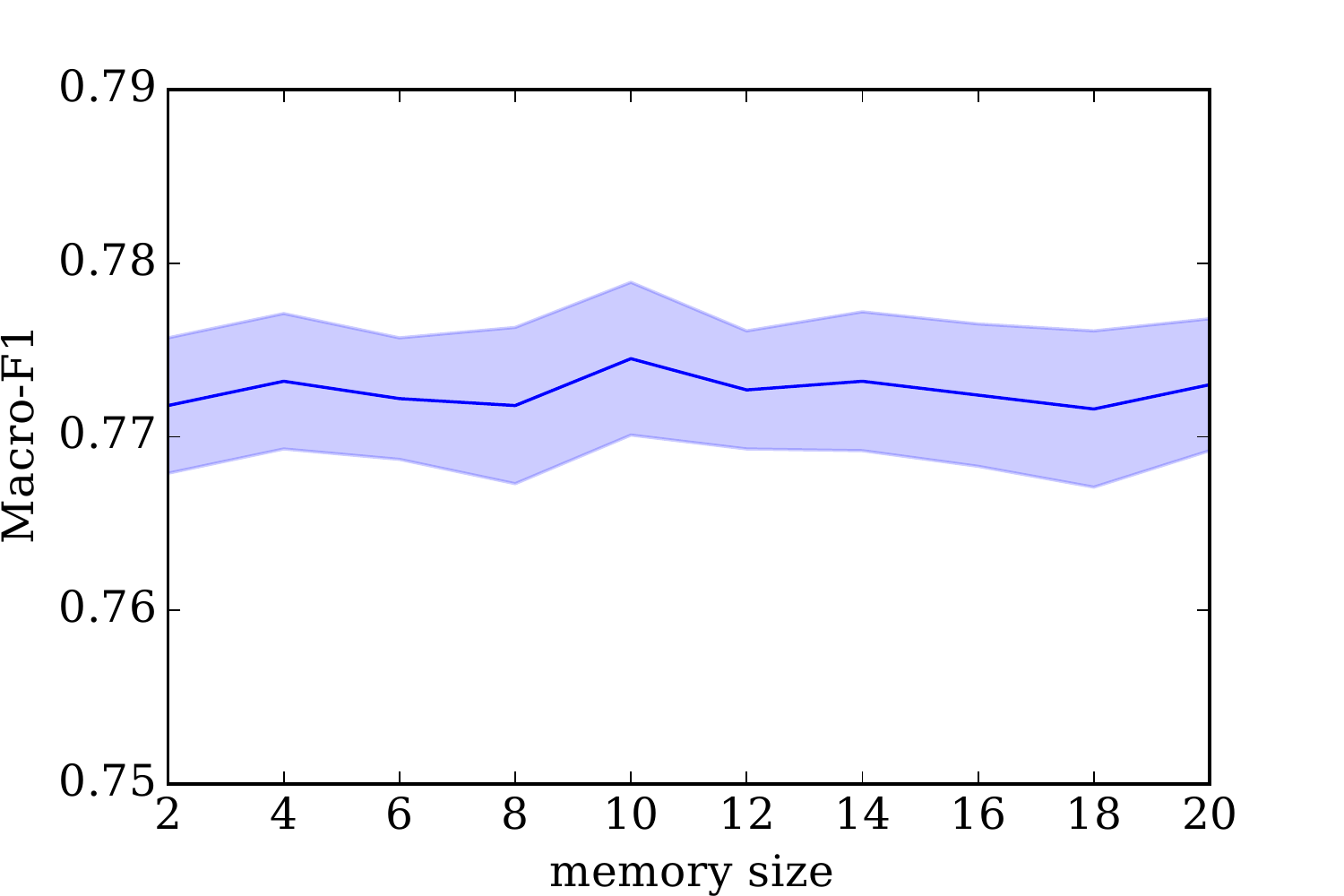}\\
(b) {\sc Dblp}
\end{minipage}
\begin{minipage}[t]{0.32\textwidth}
\centering
\includegraphics[width=1\textwidth]{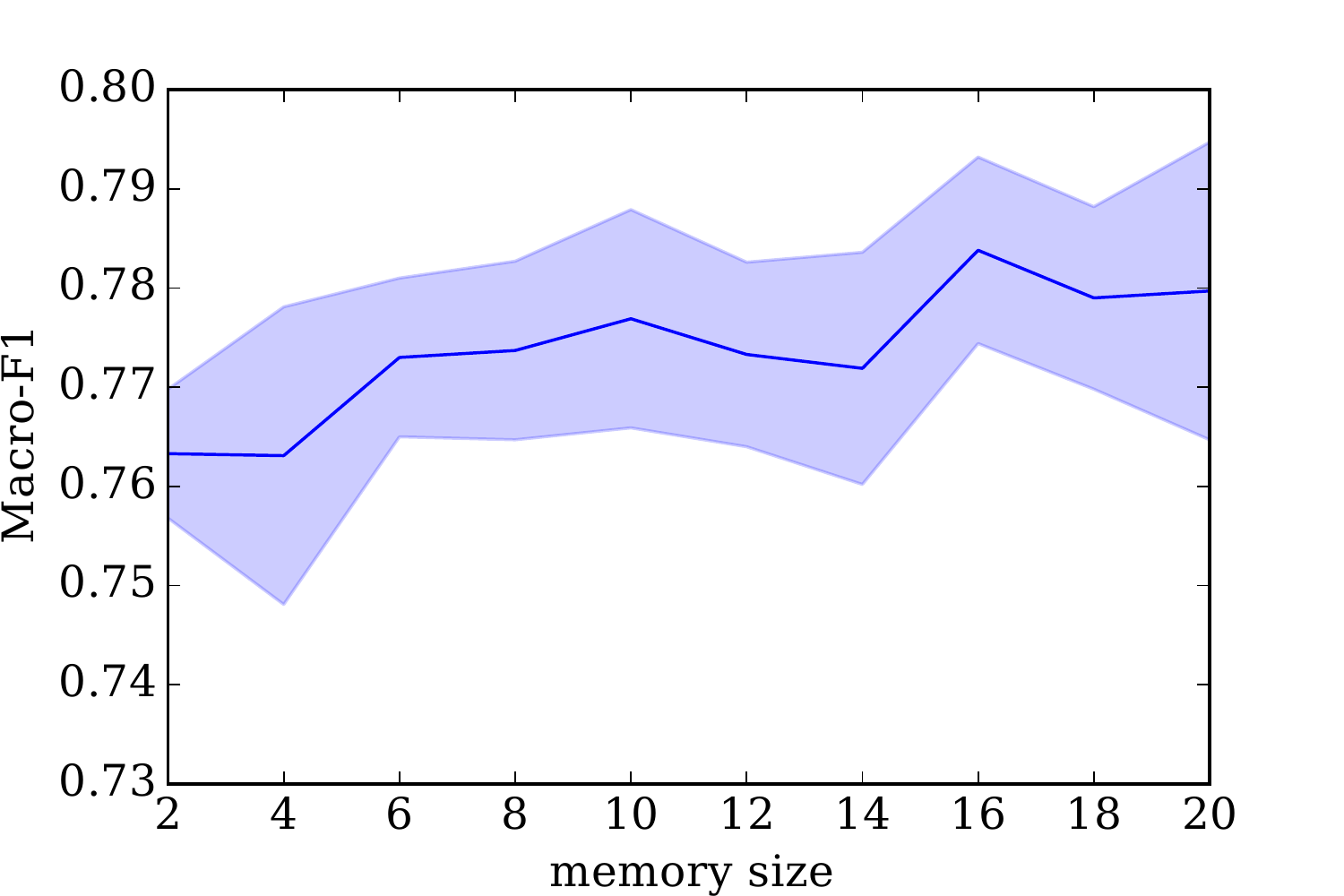}\\
(c) {\sc Twitter}
\end{minipage}
\caption{Classification performance w.r.t.  memory size (ExpaNet-S, \#hops = 1). Color regions correspond to $\pm 1$ standard deviation around the mean.  }
\label{fig:memsize}
\end{minipage}
\end{figure*}

We observe that on both {\sc Dblp} and {\sc Twitter} data sets, the performance are not sensitive to the memory size. However, on {\sc Wikipedia} data set, smaller memory size leads to slightly better performance. The reason is that on {\sc Wikipedia} data set, there is only one or very few relevant documents about a Wikipedia title, which is retrieved by the retrieval module. Adding more documents into the memory introduces more noise. However, this may not hold in real-world data: Web search usually returns many relevant documents.

\subsection{Attention Interpretation}
The attention mechanism in the ExpaNet essentially computes similarity between a short text and each document in memory. We are interested in the attention mechanism learned by our algorithm. We are interested in the following questions: which memory cells does the model learn to pay more attention to? With more number of hops, how does the attention change? Note that when the long documents are loaded into memory, their retrieval ranks are reserved: the highest ranked document is placed in Cell 1, the lowest in Cell 20. The memory networks, on the other hand, treat the memory as ``a bag of cells'' without considering the order.

\begin{figure}[h!]
\begin{minipage}[t]{0.5\textwidth}
\begin{minipage}[t]{0.5\textwidth}
\centering
\includegraphics[width=1\textwidth]{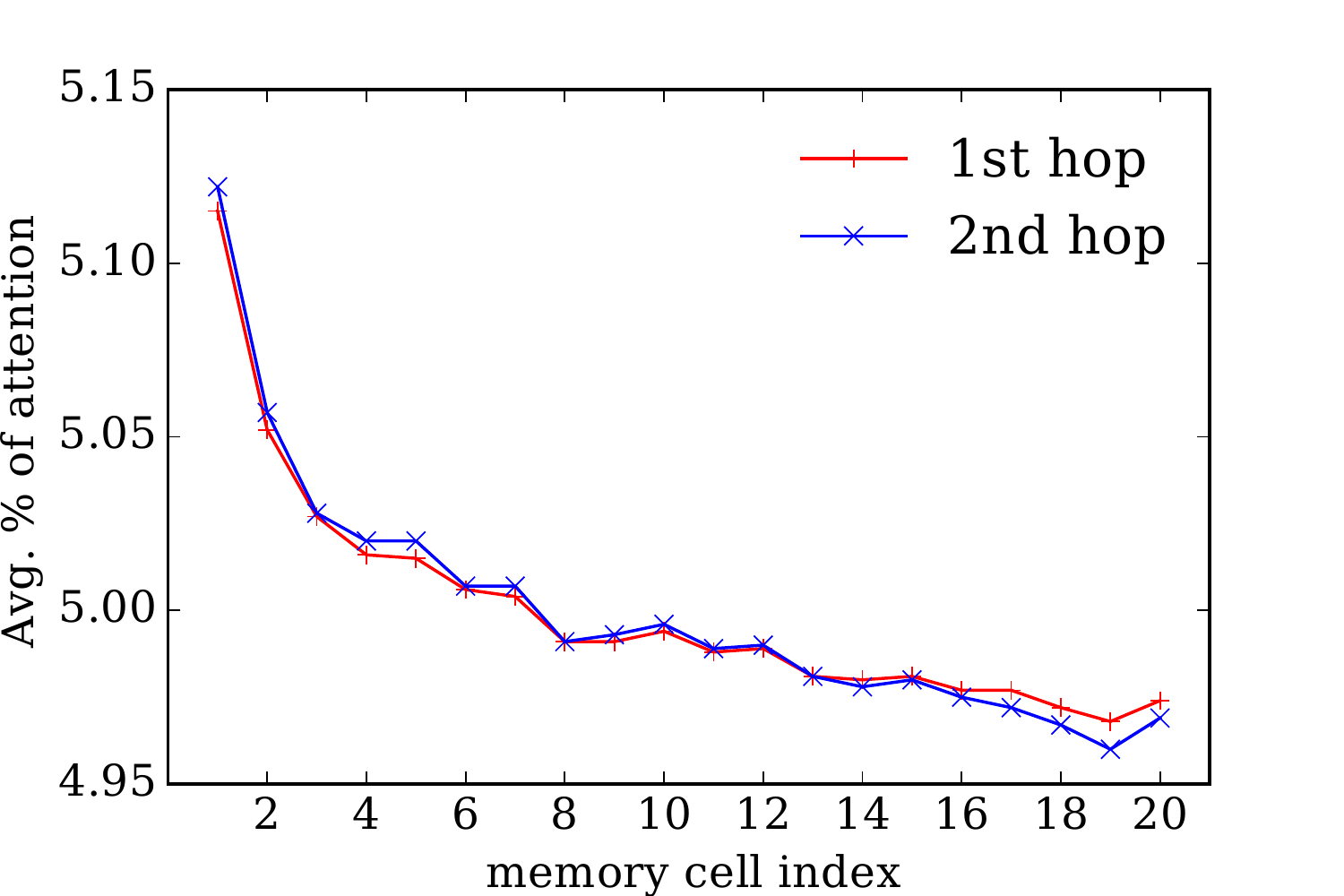}\\
(a) {\sc Wikipedia}
\end{minipage}
\begin{minipage}[t]{0.5\textwidth}
\centering
\includegraphics[width=1\textwidth]{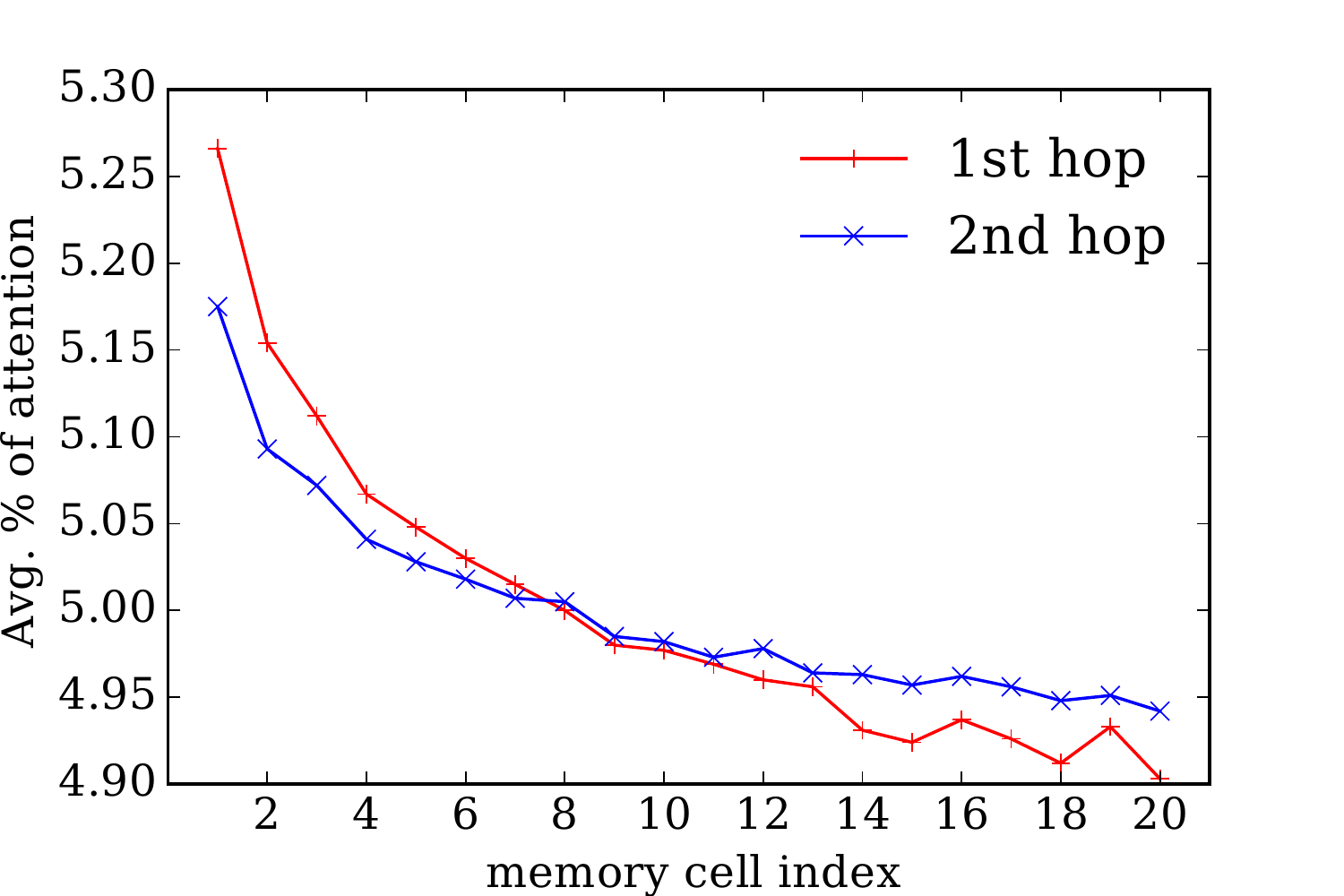}\\
(b) {\sc Dblp}
\end{minipage}
\caption{Attention distribution over memory cells. The plot of {\sc Twitter} data set is similar to {\sc Dblp} hence omitted. }
\label{fig:attn}
\end{minipage}
\end{figure}

In Figure \ref{fig:attn}, we plot the attention distribution over 20 memory cells in a soft attention memory network. The distribution is estimated by averaging the attentions on test data set. 
Hard attention memory networks have similar attention distribution but with higher variance because of its stochastic nature.
The attention distribution agrees with our prior knowledge of retrieval relevance. Memory cells from left to right hold documents with decreasing relevance scores. After training, the algorithm learns to pay more attention to cells on the left, which agrees with the relevance ranking. With more hops, attention on {\sc Dblp} and {\sc Twitter} tends to distribute uniformly across memory cells, indicating the expansion process can identify more relevant documents from the memory with more hops.



\section{Conclusion}
In this paper, we proposed an end-to-end deep memory network approach for short text expansion with a large corpus of long documents. Inspired by the human search strategy, the memory network learns to select relevant documents with attention mechanism, combine short text and expanded documents with a gating mechanism, and is trained end-to-end with short text classification as the objective. Extensive experiments on several real-world data sets show that our model significantly outperforms classical query expansion methods and methods without using external data. In the future, we plan to study how to automatically infer the optimal number of expansion hops for each short text.
\\

\noindent
\textbf{Acknowledgements.}
We thank the anonymous reviewers for their constructive comments. This work is supported by the National Institutes of Health under grant 
NLM 2R01LM010681-05 and the National Science Foundation under grant IIS-1054199. 

\bibliographystyle{ACM-Reference-Format}
\bibliography{sigproc} 


\begin{thebibliography}{00}


\ifx \showCODEN    \undefined \def \showCODEN     #1{\unskip}     \fi
\ifx \showDOI      \undefined \def \showDOI       #1{{\tt DOI:}\penalty0{#1}\ }
  \fi
\ifx \showISBNx    \undefined \def \showISBNx     #1{\unskip}     \fi
\ifx \showISBNxiii \undefined \def \showISBNxiii  #1{\unskip}     \fi
\ifx \showISSN     \undefined \def \showISSN      #1{\unskip}     \fi
\ifx \showLCCN     \undefined \def \showLCCN      #1{\unskip}     \fi
\ifx \shownote     \undefined \def \shownote      #1{#1}          \fi
\ifx \showarticletitle \undefined \def \showarticletitle #1{#1}   \fi
\ifx \showURL      \undefined \def \showURL       #1{#1}          \fi
\providecommand\bibfield[2]{#2}
\providecommand\bibinfo[2]{#2}
\providecommand\natexlab[1]{#1}
\providecommand\showeprint[2][]{arXiv:#2}

\bibitem[\protect\citeauthoryear{Bendersky and Croft}{Bendersky and
  Croft}{2009}]%
        {bendersky2009analysis}
\bibfield{author}{\bibinfo{person}{Michael Bendersky} {and}
  \bibinfo{person}{W~Bruce Croft}.} \bibinfo{year}{2009}\natexlab{}.
\newblock \showarticletitle{Analysis of long queries in a large scale search
  log}. In \bibinfo{booktitle}{{\em Proceedings of 2009 Workshop on Web Search
  Click Data}}. \bibinfo{pages}{8--14}.
\newblock


\bibitem[\protect\citeauthoryear{Buckley, Salton, Allan, and Singhal}{Buckley
  et~al\mbox{.}}{1995}]%
        {buckley1995automatic}
\bibfield{author}{\bibinfo{person}{Chris Buckley}, \bibinfo{person}{Gerard
  Salton}, \bibinfo{person}{James Allan}, {and} \bibinfo{person}{Amit
  Singhal}.} \bibinfo{year}{1995}\natexlab{}.
\newblock \showarticletitle{Automatic query expansion using SMART: TREC 3}.
\newblock \bibinfo{journal}{{\em NIST Special Publication\/}}
  (\bibinfo{year}{1995}), \bibinfo{pages}{69--69}.
\newblock


\bibitem[\protect\citeauthoryear{Cho, Van~Merri{\"e}nboer, Gulcehre, Bahdanau,
  Bougares, Schwenk, and Bengio}{Cho et~al\mbox{.}}{2014}]%
        {cho2014learning}
\bibfield{author}{\bibinfo{person}{Kyunghyun Cho}, \bibinfo{person}{Bart
  Van~Merri{\"e}nboer}, \bibinfo{person}{Caglar Gulcehre},
  \bibinfo{person}{Dzmitry Bahdanau}, \bibinfo{person}{Fethi Bougares},
  \bibinfo{person}{Holger Schwenk}, {and} \bibinfo{person}{Yoshua Bengio}.}
  \bibinfo{year}{2014}\natexlab{}.
\newblock \showarticletitle{Learning phrase representations using RNN
  encoder-decoder for statistical machine translation}.
\newblock \bibinfo{journal}{{\em arXiv preprint arXiv:1406.1078\/}}
  (\bibinfo{year}{2014}).
\newblock


\bibitem[\protect\citeauthoryear{Dumais, Banko, Brill, Lin, and Ng}{Dumais
  et~al\mbox{.}}{2002}]%
        {dumais2002web}
\bibfield{author}{\bibinfo{person}{Susan Dumais}, \bibinfo{person}{Michele
  Banko}, \bibinfo{person}{Eric Brill}, \bibinfo{person}{Jimmy Lin}, {and}
  \bibinfo{person}{Andrew Ng}.} \bibinfo{year}{2002}\natexlab{}.
\newblock \showarticletitle{Web question answering: Is more always better?}. In
  \bibinfo{booktitle}{{\em Proceedings of 25th International ACM SIGIR
  Conference on Research and Development in Information Retrieval}}.
  \bibinfo{pages}{291--298}.
\newblock


\bibitem[\protect\citeauthoryear{Efron, Organisciak, and Fenlon}{Efron
  et~al\mbox{.}}{2012}]%
        {efron2012improving}
\bibfield{author}{\bibinfo{person}{Miles Efron}, \bibinfo{person}{Peter
  Organisciak}, {and} \bibinfo{person}{Katrina Fenlon}.}
  \bibinfo{year}{2012}\natexlab{}.
\newblock \showarticletitle{Improving retrieval of short texts through document
  expansion}. In \bibinfo{booktitle}{{\em Proceedings of 35th International ACM
  SIGIR Conference on Research and Development in Information Retrieval}}.
  \bibinfo{pages}{911--920}.
\newblock


\bibitem[\protect\citeauthoryear{Fan, Chang, Hsieh, Wang, and Lin}{Fan
  et~al\mbox{.}}{2008}]%
        {fan2008liblinear}
\bibfield{author}{\bibinfo{person}{Rong-En Fan}, \bibinfo{person}{Kai-Wei
  Chang}, \bibinfo{person}{Cho-Jui Hsieh}, \bibinfo{person}{Xiang-Rui Wang},
  {and} \bibinfo{person}{Chih-Jen Lin}.} \bibinfo{year}{2008}\natexlab{}.
\newblock \showarticletitle{LIBLINEAR: A library for large linear
  classification}.
\newblock \bibinfo{journal}{{\em Journal of Machine Learning Research\/}}
  \bibinfo{volume}{9}, \bibinfo{number}{Aug} (\bibinfo{year}{2008}),
  \bibinfo{pages}{1871--1874}.
\newblock


\bibitem[\protect\citeauthoryear{Gabrilovich and Markovitch}{Gabrilovich and
  Markovitch}{2007}]%
        {gabrilovich2007computing}
\bibfield{author}{\bibinfo{person}{Evgeniy Gabrilovich} {and}
  \bibinfo{person}{Shaul Markovitch}.} \bibinfo{year}{2007}\natexlab{}.
\newblock \showarticletitle{Computing semantic relatedness using
  wikipedia-based explicit semantic analysis.}. In \bibinfo{booktitle}{{\em
  IJcAI}}, Vol.~\bibinfo{volume}{7}. \bibinfo{pages}{1606--1611}.
\newblock


\bibitem[\protect\citeauthoryear{Graves}{Graves}{2012}]%
        {graves2012supervised}
\bibfield{author}{\bibinfo{person}{Alex Graves}.}
  \bibinfo{year}{2012}\natexlab{}.
\newblock \showarticletitle{Supervised sequence labeling}.
\newblock In \bibinfo{booktitle}{{\em Supervised Sequence Labeling with
  Recurrent Neural Networks}}. \bibinfo{publisher}{Springer},
  \bibinfo{pages}{5--13}.
\newblock


\bibitem[\protect\citeauthoryear{Graves, Wayne, and Danihelka}{Graves
  et~al\mbox{.}}{2014}]%
        {graves2014neural}
\bibfield{author}{\bibinfo{person}{Alex Graves}, \bibinfo{person}{Greg Wayne},
  {and} \bibinfo{person}{Ivo Danihelka}.} \bibinfo{year}{2014}\natexlab{}.
\newblock \showarticletitle{Neural turing machines}.
\newblock \bibinfo{journal}{{\em arXiv preprint arXiv:1410.5401\/}}
  (\bibinfo{year}{2014}).
\newblock


\bibitem[\protect\citeauthoryear{Hu, Sun, Zhang, and Chua}{Hu
  et~al\mbox{.}}{2009}]%
        {hu2009exploiting}
\bibfield{author}{\bibinfo{person}{Xia Hu}, \bibinfo{person}{Nan Sun},
  \bibinfo{person}{Chao Zhang}, {and} \bibinfo{person}{Tat-Seng Chua}.}
  \bibinfo{year}{2009}\natexlab{}.
\newblock \showarticletitle{Exploiting internal and external semantics for the
  clustering of short texts using world knowledge}. In \bibinfo{booktitle}{{\em
  Proceedings of 18th ACM Conference on Information and Knowledge Management}}.
  \bibinfo{pages}{919--928}.
\newblock


\bibitem[\protect\citeauthoryear{Jang, Gu, and Poole}{Jang
  et~al\mbox{.}}{2016}]%
        {jang2016categorical}
\bibfield{author}{\bibinfo{person}{Eric Jang}, \bibinfo{person}{Shixiang Gu},
  {and} \bibinfo{person}{Ben Poole}.} \bibinfo{year}{2016}\natexlab{}.
\newblock \showarticletitle{Categorical Reparameterization with
  Gumbel-Softmax}.
\newblock \bibinfo{journal}{{\em arXiv preprint arXiv:1611.01144\/}}
  (\bibinfo{year}{2016}).
\newblock


\bibitem[\protect\citeauthoryear{Joulin, Grave, Bojanowski, and Mikolov}{Joulin
  et~al\mbox{.}}{2016}]%
        {joulin2016bag}
\bibfield{author}{\bibinfo{person}{Armand Joulin}, \bibinfo{person}{Edouard
  Grave}, \bibinfo{person}{Piotr Bojanowski}, {and} \bibinfo{person}{Tomas
  Mikolov}.} \bibinfo{year}{2016}\natexlab{}.
\newblock \showarticletitle{Bag of tricks for efficient text classification}.
\newblock \bibinfo{journal}{{\em arXiv preprint arXiv:1607.01759\/}}
  (\bibinfo{year}{2016}).
\newblock


\bibitem[\protect\citeauthoryear{Kim}{Kim}{2014}]%
        {kim2014convolutional}
\bibfield{author}{\bibinfo{person}{Yoon Kim}.} \bibinfo{year}{2014}\natexlab{}.
\newblock \showarticletitle{Convolutional neural networks for sentence
  classification}.
\newblock \bibinfo{journal}{{\em arXiv preprint arXiv:1408.5882\/}}
  (\bibinfo{year}{2014}).
\newblock


\bibitem[\protect\citeauthoryear{Kingma and Ba}{Kingma and Ba}{2014}]%
        {kingma2014adam}
\bibfield{author}{\bibinfo{person}{Diederik Kingma} {and}
  \bibinfo{person}{Jimmy Ba}.} \bibinfo{year}{2014}\natexlab{}.
\newblock \showarticletitle{Adam: A method for stochastic optimization}.
\newblock \bibinfo{journal}{{\em arXiv preprint arXiv:1412.6980\/}}
  (\bibinfo{year}{2014}).
\newblock


\bibitem[\protect\citeauthoryear{Kumar, Irsoy, Su, Bradbury, English, Pierce,
  Ondruska, Gulrajani, and Socher}{Kumar et~al\mbox{.}}{2015}]%
        {kumar2015ask}
\bibfield{author}{\bibinfo{person}{Ankit Kumar}, \bibinfo{person}{Ozan Irsoy},
  \bibinfo{person}{Jonathan Su}, \bibinfo{person}{James Bradbury},
  \bibinfo{person}{Robert English}, \bibinfo{person}{Brian Pierce},
  \bibinfo{person}{Peter Ondruska}, \bibinfo{person}{Ishaan Gulrajani}, {and}
  \bibinfo{person}{Richard Socher}.} \bibinfo{year}{2015}\natexlab{}.
\newblock \showarticletitle{Ask me anything: Dynamic memory networks for
  natural language processing}.
\newblock \bibinfo{journal}{{\em arXiv preprint arXiv:1506.07285\/}}
  (\bibinfo{year}{2015}).
\newblock


\bibitem[\protect\citeauthoryear{Le and Mikolov}{Le and Mikolov}{2014}]%
        {le2014distributed}
\bibfield{author}{\bibinfo{person}{Quoc~V Le} {and} \bibinfo{person}{Tomas
  Mikolov}.} \bibinfo{year}{2014}\natexlab{}.
\newblock \showarticletitle{Distributed Representations of Sentences and
  Documents.}. In \bibinfo{booktitle}{{\em ICML}}, Vol.~\bibinfo{volume}{14}.
  \bibinfo{pages}{1188--1196}.
\newblock


\bibitem[\protect\citeauthoryear{Lundquist, Grossman, and Frieder}{Lundquist
  et~al\mbox{.}}{1997}]%
        {lundquist1997improving}
\bibfield{author}{\bibinfo{person}{Carol Lundquist}, \bibinfo{person}{David~A
  Grossman}, {and} \bibinfo{person}{Ophir Frieder}.}
  \bibinfo{year}{1997}\natexlab{}.
\newblock \showarticletitle{Improving relevance feedback in the vector space
  model}. In \bibinfo{booktitle}{{\em Proceedings of 6th International
  Conference on Information and Knowledge Management}}. ACM,
  \bibinfo{pages}{16--23}.
\newblock


\bibitem[\protect\citeauthoryear{Macdonald and Ounis}{Macdonald and
  Ounis}{2007}]%
        {macdonald2007expertise}
\bibfield{author}{\bibinfo{person}{Craig Macdonald} {and} \bibinfo{person}{Iadh
  Ounis}.} \bibinfo{year}{2007}\natexlab{}.
\newblock \showarticletitle{Expertise drift and query expansion in expert
  search}. In \bibinfo{booktitle}{{\em Proceedings of 16th ACM Conference on
  Information and Knowledge Management}}. \bibinfo{pages}{341--350}.
\newblock


\bibitem[\protect\citeauthoryear{Merler, Galleguillos, and Belongie}{Merler
  et~al\mbox{.}}{}]%
        {merler2007recognizing}
\bibfield{author}{\bibinfo{person}{Michele Merler}, \bibinfo{person}{Carolina
  Galleguillos}, {and} \bibinfo{person}{Serge Belongie}.}
\newblock \showarticletitle{Recognizing groceries in situ using in vitro
  training data}. In \bibinfo{booktitle}{{\em CVPR 2007}}.
\newblock


\bibitem[\protect\citeauthoryear{Mikolov, Chen, Corrado, and Dean}{Mikolov
  et~al\mbox{.}}{}]%
        {mikolov2013efficient}
\bibfield{author}{\bibinfo{person}{Tomas Mikolov}, \bibinfo{person}{Kai Chen},
  \bibinfo{person}{Greg Corrado}, {and} \bibinfo{person}{Jeffrey Dean}.}
\newblock \showarticletitle{Efficient estimation of word representations in
  vector space}.
\newblock \bibinfo{journal}{{\em arXiv preprint arXiv:1301.3781\/}}
  (\bibinfo{year}{????}).
\newblock


\bibitem[\protect\citeauthoryear{Mikolov, Sutskever, Chen, Corrado, and
  Dean}{Mikolov et~al\mbox{.}}{2013}]%
        {mikolov2013distributed}
\bibfield{author}{\bibinfo{person}{Tomas Mikolov}, \bibinfo{person}{Ilya
  Sutskever}, \bibinfo{person}{Kai Chen}, \bibinfo{person}{Greg~S Corrado},
  {and} \bibinfo{person}{Jeff Dean}.} \bibinfo{year}{2013}\natexlab{}.
\newblock \showarticletitle{Distributed representations of words and phrases
  and their compositionality}. In \bibinfo{booktitle}{{\em Advances in Neural
  Information Processing Systems}}. \bibinfo{pages}{3111--3119}.
\newblock


\bibitem[\protect\citeauthoryear{Mnih, Heess, Graves, and Kavukcuoglu}{Mnih
  et~al\mbox{.}}{2014}]%
        {mnih2014recurrent}
\bibfield{author}{\bibinfo{person}{Volodymyr Mnih}, \bibinfo{person}{Nicolas
  Heess}, \bibinfo{person}{Alex Graves}, {and} \bibinfo{person}{Koray
  Kavukcuoglu}.} \bibinfo{year}{2014}\natexlab{}.
\newblock \showarticletitle{Recurrent models of visual attention}. In
  \bibinfo{booktitle}{{\em Advances in Neural Information Processing Systems}}.
  \bibinfo{pages}{2204--2212}.
\newblock


\bibitem[\protect\citeauthoryear{Munkhdalai and Yu}{Munkhdalai and Yu}{2016}]%
        {munkhdalai2016neural}
\bibfield{author}{\bibinfo{person}{Tsendsuren Munkhdalai} {and}
  \bibinfo{person}{Hong Yu}.} \bibinfo{year}{2016}\natexlab{}.
\newblock \showarticletitle{Neural Semantic Encoders}.
\newblock \bibinfo{journal}{{\em arXiv preprint arXiv:1607.04315\/}}
  (\bibinfo{year}{2016}).
\newblock


\bibitem[\protect\citeauthoryear{Palangi, Deng, Shen, Gao, He, Chen, Song, and
  Ward}{Palangi et~al\mbox{.}}{2016}]%
        {palangi2016deep}
\bibfield{author}{\bibinfo{person}{Hamid Palangi}, \bibinfo{person}{Li Deng},
  \bibinfo{person}{Yelong Shen}, \bibinfo{person}{Jianfeng Gao},
  \bibinfo{person}{Xiaodong He}, \bibinfo{person}{Jianshu Chen},
  \bibinfo{person}{Xinying Song}, {and} \bibinfo{person}{Rabab Ward}.}
  \bibinfo{year}{2016}\natexlab{}.
\newblock \showarticletitle{Deep sentence embedding using long short-term
  memory networks: Analysis and application to information retrieval}.
\newblock \bibinfo{journal}{{\em IEEE/ACM Transactions on Audio, Speech and
  Language Processing (TASLP)\/}} \bibinfo{volume}{24}, \bibinfo{number}{4}
  (\bibinfo{year}{2016}), \bibinfo{pages}{694--707}.
\newblock


\bibitem[\protect\citeauthoryear{Phan, Nguyen, and Horiguchi}{Phan
  et~al\mbox{.}}{2008}]%
        {phan2008learning}
\bibfield{author}{\bibinfo{person}{Xuan-Hieu Phan}, \bibinfo{person}{Le-Minh
  Nguyen}, {and} \bibinfo{person}{Susumu Horiguchi}.}
  \bibinfo{year}{2008}\natexlab{}.
\newblock \showarticletitle{Learning to classify short and sparse text \& web
  with hidden topics from large-scale data collections}. In
  \bibinfo{booktitle}{{\em Proceedings of 17th International Conference on
  World Wide Web}}. ACM, \bibinfo{pages}{91--100}.
\newblock


\bibitem[\protect\citeauthoryear{Rocchio}{Rocchio}{1971}]%
        {rocchio1971relevance}
\bibfield{author}{\bibinfo{person}{Joseph~John Rocchio}.}
  \bibinfo{year}{1971}\natexlab{}.
\newblock \showarticletitle{Relevance feedback in information retrieval}.
\newblock  (\bibinfo{year}{1971}).
\newblock


\bibitem[\protect\citeauthoryear{Sahami and Heilman}{Sahami and
  Heilman}{2006}]%
        {sahami2006web}
\bibfield{author}{\bibinfo{person}{Mehran Sahami} {and}
  \bibinfo{person}{Timothy~D Heilman}.} \bibinfo{year}{2006}\natexlab{}.
\newblock \showarticletitle{A web-based kernel function for measuring the
  similarity of short text snippets}. In \bibinfo{booktitle}{{\em Proceedings
  of 15th International Conference on World Wide Web}}. ACM,
  \bibinfo{pages}{377--386}.
\newblock


\bibitem[\protect\citeauthoryear{Schlaefer, Chu-Carroll, Nyberg, Fan, Zadrozny,
  and Ferrucci}{Schlaefer et~al\mbox{.}}{2011}]%
        {schlaefer2011statistical}
\bibfield{author}{\bibinfo{person}{Nico Schlaefer}, \bibinfo{person}{Jennifer
  Chu-Carroll}, \bibinfo{person}{Eric Nyberg}, \bibinfo{person}{James Fan},
  \bibinfo{person}{Wlodek Zadrozny}, {and} \bibinfo{person}{David Ferrucci}.}
  \bibinfo{year}{2011}\natexlab{}.
\newblock \showarticletitle{Statistical source expansion for question
  answering}. In \bibinfo{booktitle}{{\em Proceedings of 20th ACM International
  Conference on Information and Knowledge Management}}.
  \bibinfo{pages}{345--354}.
\newblock


\bibitem[\protect\citeauthoryear{Sukhbaatar, Weston, Fergus,
  et~al\mbox{.}}{Sukhbaatar et~al\mbox{.}}{2015}]%
        {sukhbaatar2015end}
\bibfield{author}{\bibinfo{person}{Sainbayar Sukhbaatar},
  \bibinfo{person}{Jason Weston}, \bibinfo{person}{Rob Fergus}, {and}
  \bibinfo{person}{others}.} \bibinfo{year}{2015}\natexlab{}.
\newblock \showarticletitle{End-to-end memory networks}. In
  \bibinfo{booktitle}{{\em Advances in Neural Information Processing Systems}}.
  \bibinfo{pages}{2440--2448}.
\newblock


\bibitem[\protect\citeauthoryear{Tang, Qu, and Mei}{Tang
  et~al\mbox{.}}{2015a}]%
        {tang2015pte}
\bibfield{author}{\bibinfo{person}{Jian Tang}, \bibinfo{person}{Meng Qu}, {and}
  \bibinfo{person}{Qiaozhu Mei}.} \bibinfo{year}{2015}\natexlab{a}.
\newblock \showarticletitle{PTE: Predictive text embedding through large-scale
  heterogeneous text networks}. In \bibinfo{booktitle}{{\em Proceedings of 21st
  ACM SIGKDD International Conference on Knowledge Discovery and Data Mining}}.
  \bibinfo{pages}{1165--1174}.
\newblock


\bibitem[\protect\citeauthoryear{Tang, Qu, Wang, Zhang, Yan, and Mei}{Tang
  et~al\mbox{.}}{2015b}]%
        {tang2015line}
\bibfield{author}{\bibinfo{person}{Jian Tang}, \bibinfo{person}{Meng Qu},
  \bibinfo{person}{Mingzhe Wang}, \bibinfo{person}{Ming Zhang},
  \bibinfo{person}{Jun Yan}, {and} \bibinfo{person}{Qiaozhu Mei}.}
  \bibinfo{year}{2015}\natexlab{b}.
\newblock \showarticletitle{Line: Large-scale information network embedding}.
  In \bibinfo{booktitle}{{\em Proceedings of 24th International Conference on
  World Wide Web}}. ACM, \bibinfo{pages}{1067--1077}.
\newblock


\bibitem[\protect\citeauthoryear{Williams}{Williams}{1992}]%
        {williams1992simple}
\bibfield{author}{\bibinfo{person}{Ronald~J Williams}.}
  \bibinfo{year}{1992}\natexlab{}.
\newblock \showarticletitle{Simple statistical gradient-following algorithms
  for connectionist reinforcement learning}.
\newblock \bibinfo{journal}{{\em Machine learning\/}} \bibinfo{volume}{8},
  \bibinfo{number}{3-4} (\bibinfo{year}{1992}), \bibinfo{pages}{229--256}.
\newblock


\bibitem[\protect\citeauthoryear{Xu, Ba, Kiros, Cho, Courville, Salakhutdinov,
  Zemel, and Bengio}{Xu et~al\mbox{.}}{2015}]%
        {xu2015show}
\bibfield{author}{\bibinfo{person}{Kelvin Xu}, \bibinfo{person}{Jimmy Ba},
  \bibinfo{person}{Ryan Kiros}, \bibinfo{person}{Kyunghyun Cho},
  \bibinfo{person}{Aaron Courville}, \bibinfo{person}{Ruslan Salakhutdinov},
  \bibinfo{person}{Richard~S Zemel}, {and} \bibinfo{person}{Yoshua Bengio}.}
  \bibinfo{year}{2015}\natexlab{}.
\newblock \showarticletitle{Show, Attend and Tell: Neural Image Caption
  Generation with Visual Attention.}. In \bibinfo{booktitle}{{\em ICML}},
  Vol.~\bibinfo{volume}{14}. \bibinfo{pages}{77--81}.
\newblock


\bibitem[\protect\citeauthoryear{Zhai and Lafferty}{Zhai and Lafferty}{2001a}]%
        {zhai2001model}
\bibfield{author}{\bibinfo{person}{Chengxiang Zhai} {and} \bibinfo{person}{John
  Lafferty}.} \bibinfo{year}{2001}\natexlab{a}.
\newblock \showarticletitle{Model-based feedback in the language modeling
  approach to information retrieval}. In \bibinfo{booktitle}{{\em Proceedings
  of 10th International Conference on Information and Knowledge Management}}.
  ACM, \bibinfo{pages}{403--410}.
\newblock


\bibitem[\protect\citeauthoryear{Zhai and Lafferty}{Zhai and Lafferty}{2001b}]%
        {zhai2001study}
\bibfield{author}{\bibinfo{person}{Chengxiang Zhai} {and} \bibinfo{person}{John
  Lafferty}.} \bibinfo{year}{2001}\natexlab{b}.
\newblock \showarticletitle{A study of smoothing methods for language models
  applied to ad hoc information retrieval}. In \bibinfo{booktitle}{{\em
  Proceedings of 24th International ACM SIGIR Conference on Research and
  Development in Information Retrieval}}. \bibinfo{pages}{334--342}.
\newblock


\end{thebibliography}



\begin{thebibliography}{00}


\ifx \showCODEN    \undefined \def \showCODEN     #1{\unskip}     \fi
\ifx \showDOI      \undefined \def \showDOI       #1{{\tt DOI:}\penalty0{#1}\ }
  \fi
\ifx \showISBNx    \undefined \def \showISBNx     #1{\unskip}     \fi
\ifx \showISBNxiii \undefined \def \showISBNxiii  #1{\unskip}     \fi
\ifx \showISSN     \undefined \def \showISSN      #1{\unskip}     \fi
\ifx \showLCCN     \undefined \def \showLCCN      #1{\unskip}     \fi
\ifx \shownote     \undefined \def \shownote      #1{#1}          \fi
\ifx \showarticletitle \undefined \def \showarticletitle #1{#1}   \fi
\ifx \showURL      \undefined \def \showURL       #1{#1}          \fi
\providecommand\bibfield[2]{#2}
\providecommand\bibinfo[2]{#2}
\providecommand\natexlab[1]{#1}
\providecommand\showeprint[2][]{arXiv:#2}

\bibitem[\protect\citeauthoryear{American Mathematical Society}{American
  Mathematical Society}{2015}]%
        {Amsthm15}
American Mathematical Society \bibinfo{year}{2015}\natexlab{}.
\newblock \bibinfo{booktitle}{{\em Using the amsthm Package}}.
\newblock American Mathematical Society.
\newblock
\newblock
\shownote{\url{http://www.ctan.org/pkg/amsthm}.}


\bibitem[\protect\citeauthoryear{Bowman, Debray, and Peterson}{Bowman
  et~al\mbox{.}}{1993}]%
        {bowman:reasoning}
\bibfield{author}{\bibinfo{person}{Mic Bowman}, \bibinfo{person}{Saumya~K.
  Debray}, {and} \bibinfo{person}{Larry~L. Peterson}.}
  \bibinfo{year}{1993}\natexlab{}.
\newblock \showarticletitle{Reasoning About Naming Systems}.
\newblock \bibinfo{journal}{{\em ACM Trans. Program. Lang. Syst.\/}}
  \bibinfo{volume}{15}, \bibinfo{number}{5} (\bibinfo{date}{November}
  \bibinfo{year}{1993}), \bibinfo{pages}{795--825}.
\newblock
\showDOI{%
\url{http://dx.doi.org/10.1145/161468.161471}}


\bibitem[\protect\citeauthoryear{Braams}{Braams}{1991}]%
        {braams:babel}
\bibfield{author}{\bibinfo{person}{Johannes Braams}.}
  \bibinfo{year}{1991}\natexlab{}.
\newblock \showarticletitle{Babel, a Multilingual Style-Option System for Use
  with LaTeX's Standard Document Styles}.
\newblock \bibinfo{journal}{{\em TUGboat\/}} \bibinfo{volume}{12},
  \bibinfo{number}{2} (\bibinfo{date}{June} \bibinfo{year}{1991}),
  \bibinfo{pages}{291--301}.
\newblock


\bibitem[\protect\citeauthoryear{Clark}{Clark}{1991}]%
        {clark:pct}
\bibfield{author}{\bibinfo{person}{Malcolm Clark}.}
  \bibinfo{year}{1991}\natexlab{}.
\newblock \showarticletitle{Post Congress Tristesse}. In
  \bibinfo{booktitle}{{\em TeX90 Conference Proceedings}}. TeX Users Group,
  \bibinfo{pages}{84--89}.
\newblock


\bibitem[\protect\citeauthoryear{Fear}{Fear}{2005}]%
        {Fear05}
\bibfield{author}{\bibinfo{person}{Simon Fear}.}
  \bibinfo{year}{2005}\natexlab{}.
\newblock \bibinfo{booktitle}{{\em Publication quality tables in {\LaTeX}}}.
\newblock
\newblock
\shownote{\url{http://www.ctan.org/pkg/booktabs}.}


\bibitem[\protect\citeauthoryear{Herlihy}{Herlihy}{1993}]%
        {herlihy:methodology}
\bibfield{author}{\bibinfo{person}{Maurice Herlihy}.}
  \bibinfo{year}{1993}\natexlab{}.
\newblock \showarticletitle{A Methodology for Implementing Highly Concurrent
  Data Objects}.
\newblock \bibinfo{journal}{{\em ACM Trans. Program. Lang. Syst.\/}}
  \bibinfo{volume}{15}, \bibinfo{number}{5} (\bibinfo{date}{November}
  \bibinfo{year}{1993}), \bibinfo{pages}{745--770}.
\newblock
\showDOI{%
\url{http://dx.doi.org/10.1145/161468.161469}}


\bibitem[\protect\citeauthoryear{Lamport}{Lamport}{1986}]%
        {Lamport:LaTeX}
\bibfield{author}{\bibinfo{person}{Leslie Lamport}.}
  \bibinfo{year}{1986}\natexlab{}.
\newblock \bibinfo{booktitle}{{\em LaTeX User's Guide and Document Reference
  Manual}}.
\newblock \bibinfo{publisher}{Addison-Wesley Publishing Company},
  \bibinfo{address}{Reading, Massachusetts}.
\newblock


\bibitem[\protect\citeauthoryear{Salas and Hille}{Salas and Hille}{1978}]%
        {salas:calculus}
\bibfield{author}{\bibinfo{person}{S.L. Salas} {and} \bibinfo{person}{Einar
  Hille}.} \bibinfo{year}{1978}\natexlab{}.
\newblock \bibinfo{booktitle}{{\em Calculus: One and Several Variable}}.
\newblock \bibinfo{publisher}{John Wiley and Sons}, \bibinfo{address}{New
  York}.
\newblock


\end{thebibliography}

\end{document}